\documentclass[preprint,3p,times,twocolumn]{elsarticle}

\usepackage[utf8]{inputenc}

\usepackage{epsf,graphicx}

\usepackage{subcaption}
\usepackage{url}
\usepackage{hyperref}

\usepackage{amssymb}
\usepackage{latexsym}
\usepackage{amsmath}

\usepackage{textcomp}
\usepackage{enumitem}
\usepackage{bold-extra}

\usepackage[switch]{lineno}

\biboptions{numbers}

\journal{Computer Methods and Programs in Biomedicine}

\begin{document}
	
\begin{frontmatter}
	
\title{Acute and sub-acute stroke lesion segmentation from multimodal MRI}

\author{Albert~Cl\`erigues \corref{corr1}}
\author{Sergi~Valverde \corref{}}
\author{Jose~Bernal \corref{}}
\author{Jordi~Freixenet \corref{}}
\author{Arnau~Oliver \corref{}}
\author{Xavier~Llad\'o \corref{}}

\address{Institute of Computer Vision and Robotics, University of Girona, Spain}

\cortext[corr1]{Corresponding author. A. Cl\`erigues, Ed. P-IV, Campus Montilivi, University of Girona, 17003 Girona (Spain). e-mail: \mbox{albert.clerigues@udg.edu}. Phone: +34 683645681; Fax: +34 \mbox{972 418976}.}

\begin{abstract}

	
\textbf{Background and objective. } Acute stroke lesion segmentation tasks are of great clinical interest as they can help doctors make better informed time-critical treatment decisions. Magnetic resonance imaging (MRI) is time demanding but can provide images that are considered the gold standard for diagnosis. Automated stroke lesion segmentation can provide with an estimate of the location and volume of the lesioned tissue, which can help in the clinical practice to better assess and evaluate the risks of each treatment.

\textbf{Methods.} We propose a deep learning methodology for acute and sub-acute stroke lesion segmentation using multimodal MR imaging. We pre-process the data to facilitate learning features based on the symmetry of brain hemispheres. The issue of class imbalance is tackled using small patches with a balanced training patch sampling strategy and a dynamically weighted loss function. Moreover, a combination of whole patch predictions, using a U-Net based CNN architecture, and high degree of overlapping patches reduces the need for additional post-processing.

\textbf{Results. } The proposed method is evaluated using two public datasets from the 2015 Ischemic Stroke Lesion Segmentation challenge (ISLES 2015). These involve the tasks of sub-acute stroke lesion segmentation (SISS) and acute stroke penumbra estimation (SPES) from multiple diffusion, perfusion and anatomical MRI modalities. The performance is compared against state-of-the-art methods with a blind online testing set evaluation on each of the challenges. At the time of submitting this manuscript, our approach is the first method in the online rankings for the SISS (DSC=$0.59\pm0.31$) and SPES sub-tasks (DSC=$0.84\pm0.10$). When compared with the rest of submitted strategies, we achieve top rank performance with a lower Hausdorff distance.

\textbf{Conclusions.} Better segmentation results are obtained by leveraging the anatomy and pathophysiology of acute stroke lesions and using a combined approach to minimize the effects of class imbalance. The same training procedure is used for both tasks, showing the proposed methodology can generalize well enough to deal with different unrelated tasks and imaging modalities without hyper-parameter tuning. In order to promote the reproducibility of our results, a public version of the proposed method has been released to the scientific community at \url{https://github.com/NIC-VICOROB/stroke-mri-segmentation}.
\end{abstract}

	
\end{frontmatter}


\section{Introduction}

Stroke is a medical condition by which an abnormal blood flow in the brain causes the death of cerebral tissue. Stroke is the third cause of morbidity worldwide, after myocardial infarction and cancer, and the most prevalent cause of acquired disability \cite{Redon2011}. The affected tissue in the acute phase can be divided into three concentric regions depending on the potential for recovery, also referred as salvageability: core, penumbra and benign oligemia \cite{Rekik2012}. The core, located at the center, is formed by irreversibly damaged tissue from a fatally low blood supply. The penumbra, located around the core, represents tissue at risk but that can still be recovered if blood flow is quickly restored. Finally, the benign oligemia is the outer most ring whose vascularity has been altered but is not at risk of damage. Once the symptoms of stroke have been identified, a shorter time to treatment is highly correlated with a positive outcome \cite{Sheth2015}. Mechanical thrombectomy is a strongly recommended option for eligible patients \cite{Campbell2017}. However, this surgery is not free of risks. An overall complication rate of 15.3\% was observed in a year long study \cite{Singh2017}. In the treatment decision context, an estimate of the salvageable tissue can aid physicians take more informed treatment decisions.

The Ischemic Stroke Lesion Segmentation (ISLES) challenge started in 2015 to provide a platform for fair and direct comparison of automated methods. It included two sub-tasks, the sub-acute ischemic stroke lesion segmentation (SISS) and the acute stroke penumbra estimation (SPES). The following ISLES 2016 and 2017 editions changed its focus from lesion segmentation to chronic lesion outcome prediction from MRI. In the 2015 ISLES workshop results, the top three methods in the SPES sub-task all used Random Decision Forests (RDFs) \cite{TinKamHo1995} using hand-crafted features \cite{maieo, mckir, robbd}. RDFs were typically used in methods for stroke lesion segmentation due to their excellent generalization properties, which make them well suited for difficult tasks with few training samples \cite{Maier2015}. Recent advances on convolutional neural networks (CNNs) \cite{LeCun1989} have achieved superior results and are currently replacing RDFs in most state-of-the-art methods. In contrast with RDFs, CNNs enable the joint learning of optimal features and  classification criteria at training time for the specific task. However, these kind of networks are still restricted by the architectural design, the amount and quality of available data and the training procedure. Recently, advances in regularization techniques and data imbalance handling allow for increased CNN generalization performance in brain lesion segmentation that rivals that of RDFs. The best method in the SISS sub-task of the 2015 ISLES workshop employed a deep learning strategy consisting of a dual path encoder network with a conditional random field (CRF) post-processing \cite{kamnk1}. More recently, Zhang et al \cite{Zhang2018} achieved comparable results by using a similar CNN trained with a deep supervision technique and a multi-scale loss function.
Despite the good results of these kind of networks, the U-Net architecture \cite{Ronneberger2015}, an encoder-decoder network, is replacing other state-of-the-art architectures for stroke lesion segmentation. This is clearly seen in the submissions for the ISLES 2017 challenge, where 10 out of the 14 participating methods, including the top three, used CNNs based on the U-Net architecture \cite{Winzeck2018}.

In this work, we present a deep learning approach for acute and sub-acute stroke lesion segmentation from multimodal MRI images. We use a 3D asymmetric encoder-decoder network based on the U-Net architecture with global and local residual connections. Within our approach, the class imbalance issue is alleviated with the use of small patches with balanced training patch sampling strategies and a dynamically weighted loss function. Additionally, we pre-process the provided images to facilitate using the symmetry property of brain hemispheres. The methodology is evaluated by cross-validation with the training images and with a blind online testing set evaluation against other state-of-the-art methods. The proposed approach demonstrates state of the art performance by ranking first in the testing leaderboard of both challenges \cite{isles15website} without any dataset specific tuning.

\section{Data}

For evaluation of the proposed methodology we use the public datasets provided for the two sub-tasks of the 2015 ISLES challenge \cite{Maier2017isles}. They both encompass stroke lesion segmentation tasks from MRI imaging but using different imaging modalities and acquisition time since onset.

\subsection{SISS dataset}
For the sub-acute ischemic stroke segmentation (SISS) sub-task, a dataset was provided with 28 training and 36 testing cases acquired in the first week after onset. For each case, 4 co-registered multimodal images were provided including anatomical (T1, T2, FLAIR) and diffusion (DWI) MRI. The images were acquired as 3D volumes of $230\times230\times153$ dimensions at $1\times1\times1$ mm spacing. All four MRI modalities were used for evaluation of the proposed approach. For the training images, the provided gold standard, the whole lesion extent, was manually segmented by an experienced medical doctor. 

\subsection{SPES dataset}
The acute stroke penumbra estimation sub-task (SPES) included 30 training and 20 testing cases acquired in the first day after onset. For each case, 7 co-registered modalities were provided including anatomical (T1 contrast, T2), diffusion (DWI) and perfusion (CBF, CBV, TTP, Tmax) MRI. The images were acquired as 3D volumes of $96\times110\times71$ dimensions at $2\times2\times2$ mm spacing. All seven MRI modalities were used for evaluation of the proposed approach. For the training images, the gold standard segmentation, the penumbra label, was obtained as the mismatch between whole lesion extent and the core delineated in perfusion and diffusion images respectively.


\section{Methodology}

We propose a 3D patch based deep learning method using an asymmetrical residual CNN based on the U-Net architecture \cite{Ronneberger2015}. Within our approach, the class imbalance issue is addressed with a combination of techniques including the use of small patches ($24\times24\times16$) and a weighted loss function. We also regularize the training procedure with dropout \cite{Srivastava2014}, data augmentation and early stopping. For image segmentation, the use of whole patch predictions with a high degree of overlap minimizes the need for additional post-processing. In the following, we briefly describe the main components and implementation details of our methodology.

\subsection{Data pre-processing}
\label{subsec:preprocessing}
The given images are first pre-processed with a symmetric modality augmentation to allow learning of features based on the symmetry of brain hemispheres despite the small receptive field of the used patches. Explicit symmetry information was already shown to improve results for chronic stroke lesion segmentation \cite{Wang2016}. In our case, instead of using one patch per hemisphere in a multi-path network we use a single joint patch with a single-path network. In practice, we augment the provided modality images with symmetric versions that swap the left and right hemispheres. We first flip one of the images along the mid-sagittal axis and then we apply FSL FLIRT \cite{Jenkinson2002} to perform a linear registration between the original and flipped image. Finally, the rest of modalities are registered using the same transformation. Figure~\ref{fig:sym} shows an example of the resulting symmetrically augmented modalities. These are then appended to the provided ones, effectively doubling the number of images for each patient. In this way, a single extracted patch will also include intensity information from the opposite hemisphere.

\def\modsubfig{0.162\textwidth}
\begin{figure*}[t]
	\centering
	\begin{subfigure}[b]{\modsubfig}
		\includegraphics[width=\textwidth,trim={0px 0px 0px 0px},clip]{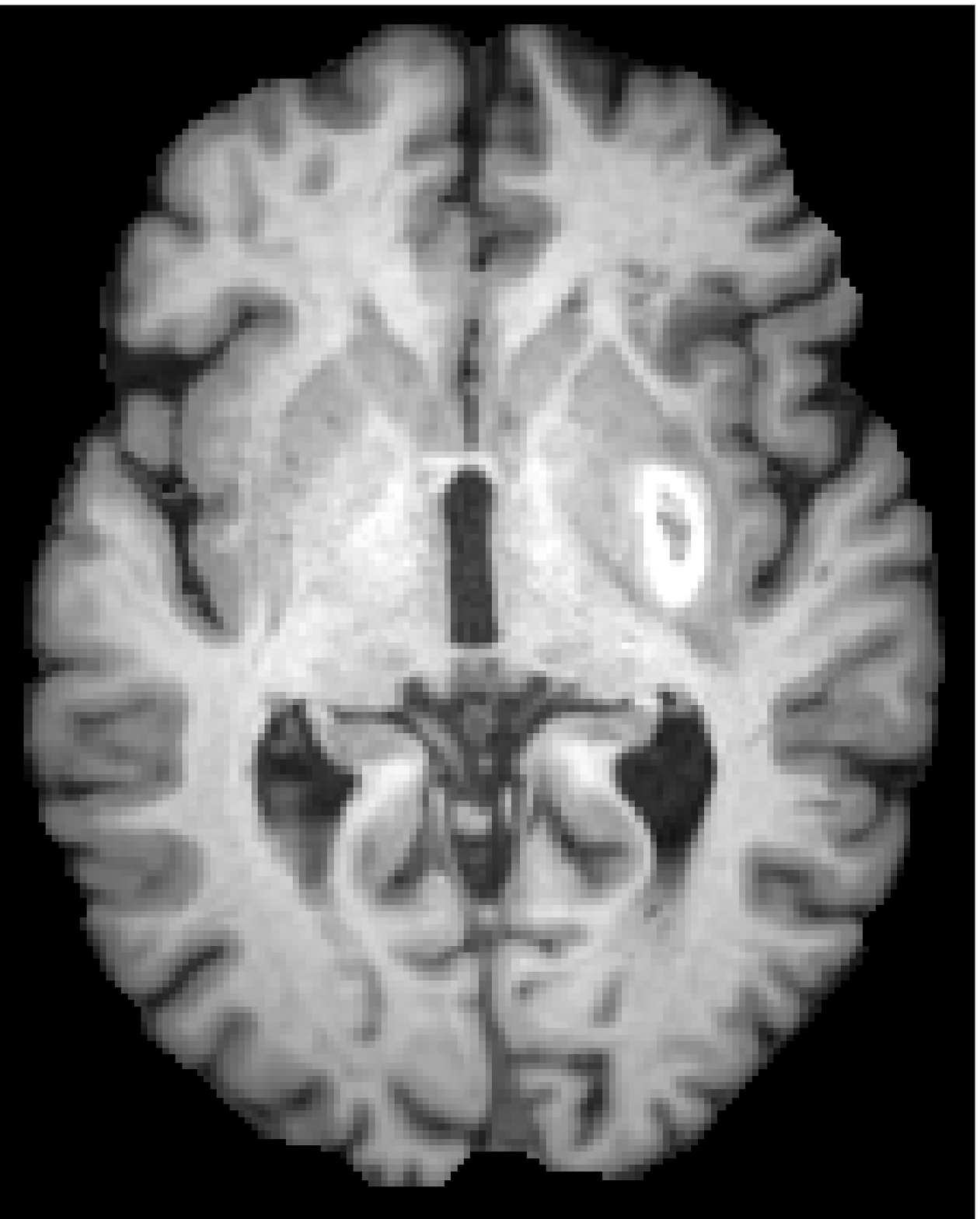}
		\caption{T1}
	\end{subfigure} 
	\begin{subfigure}[b]{\modsubfig}
		\includegraphics[width=\textwidth,trim={0px 0px 0px 0px},clip]{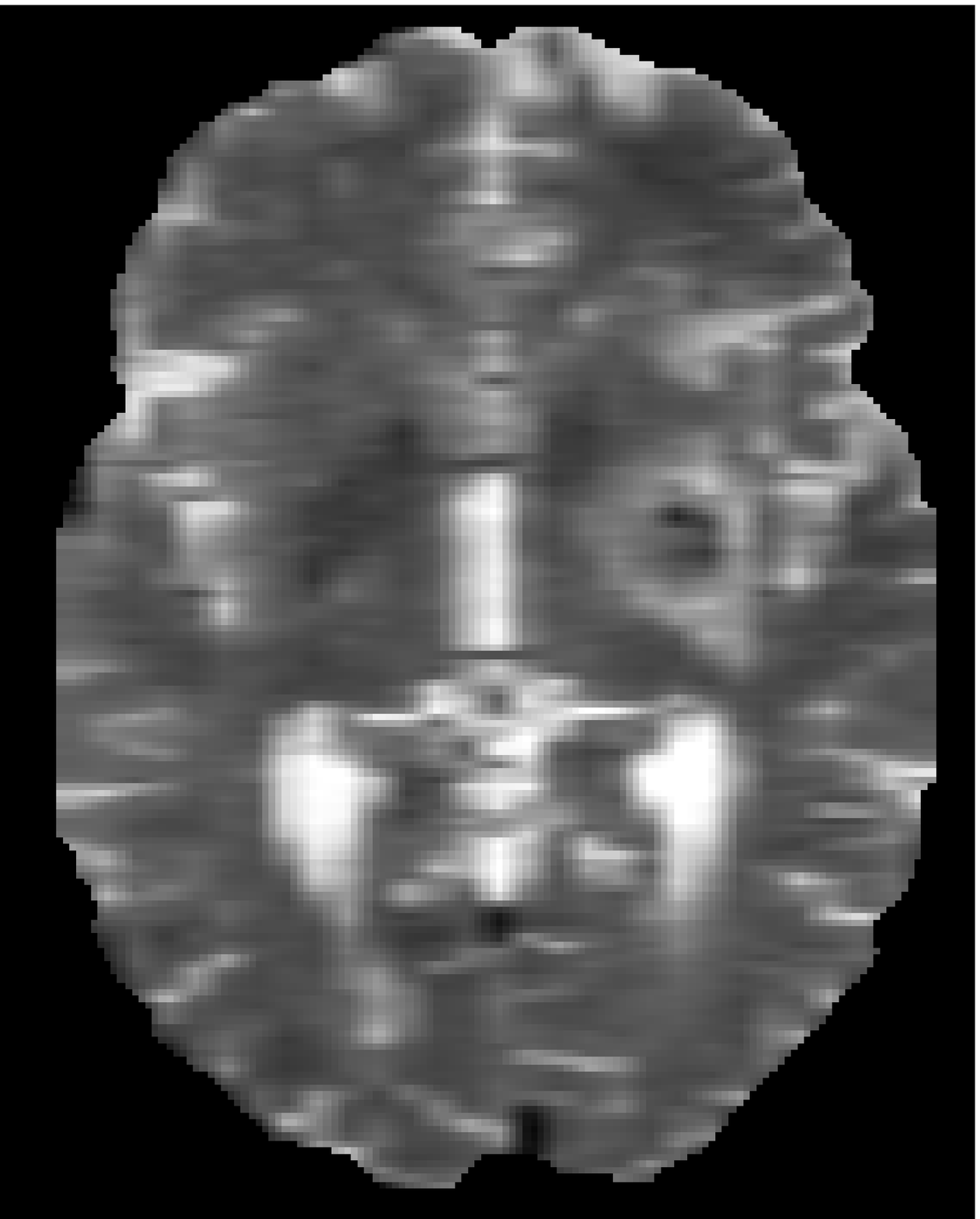}
		\caption{T2}
	\end{subfigure}
	\begin{subfigure}[b]{\modsubfig}
		\includegraphics[width=\textwidth,trim={0px 0px 0px 0px},clip]{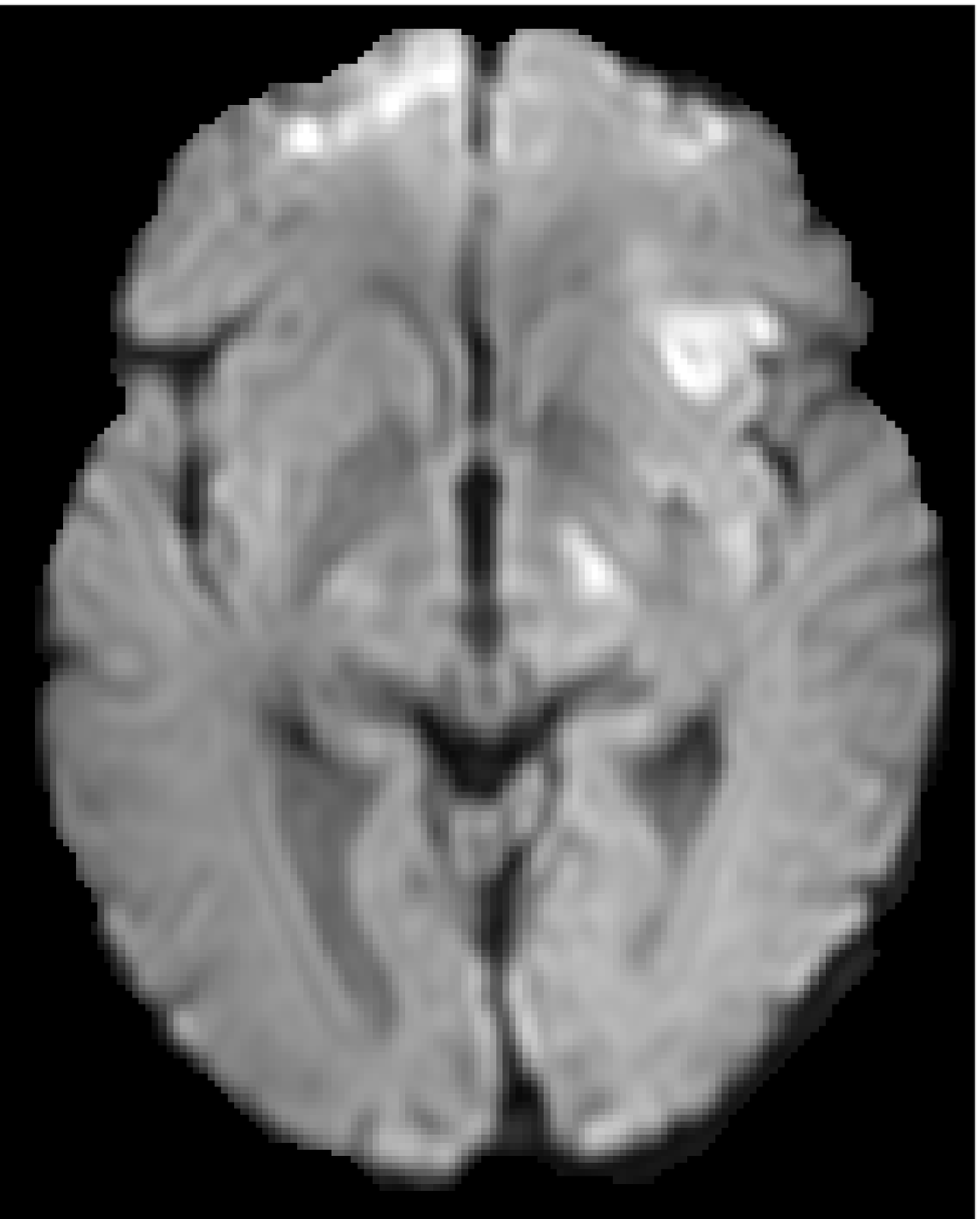}
		\caption{DWI}
	\end{subfigure} 
	\begin{subfigure}[b]{\modsubfig}
		\includegraphics[width=\textwidth,trim={0px 0px 0px 0px},clip]{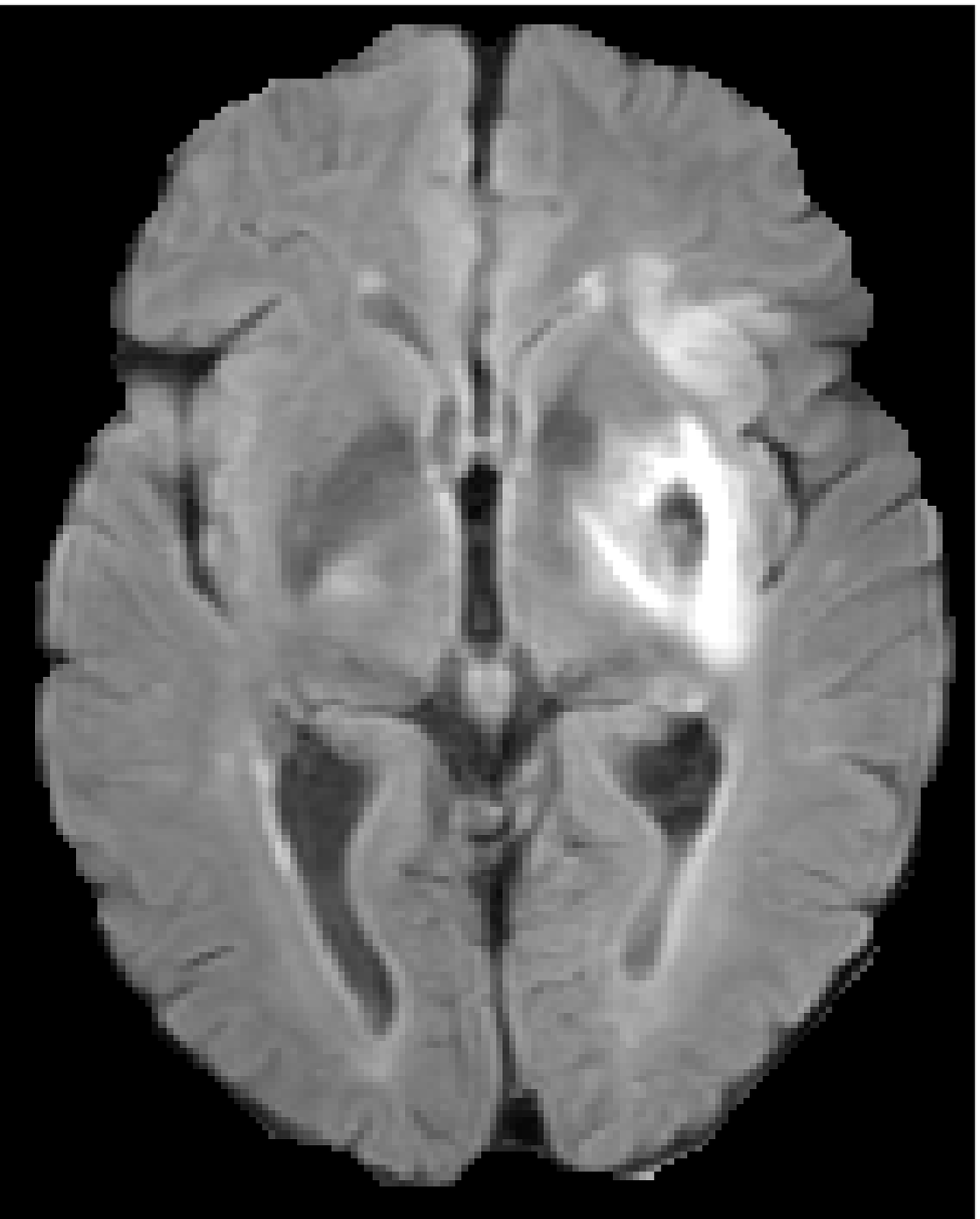}
		\caption{FLAIR}
	\end{subfigure}
	\\ 
	\begin{subfigure}[b]{\modsubfig}
		\includegraphics[width=\textwidth,trim={0px 0px 0px 0px},clip]{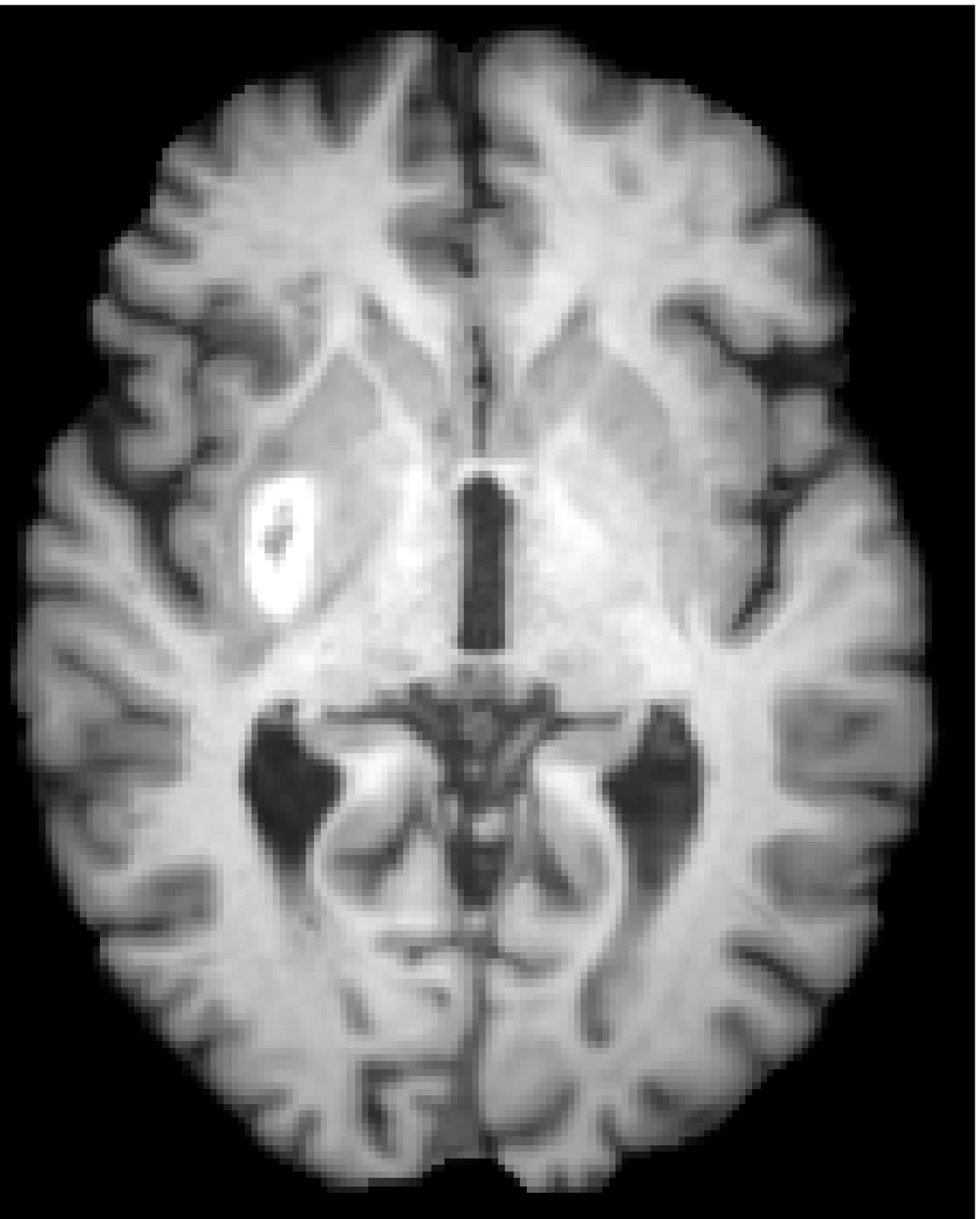}
		\caption{Sym. T1}
	\end{subfigure} 
	\begin{subfigure}[b]{\modsubfig}
		\includegraphics[width=\textwidth,trim={0px 0px 0px 0px},clip]{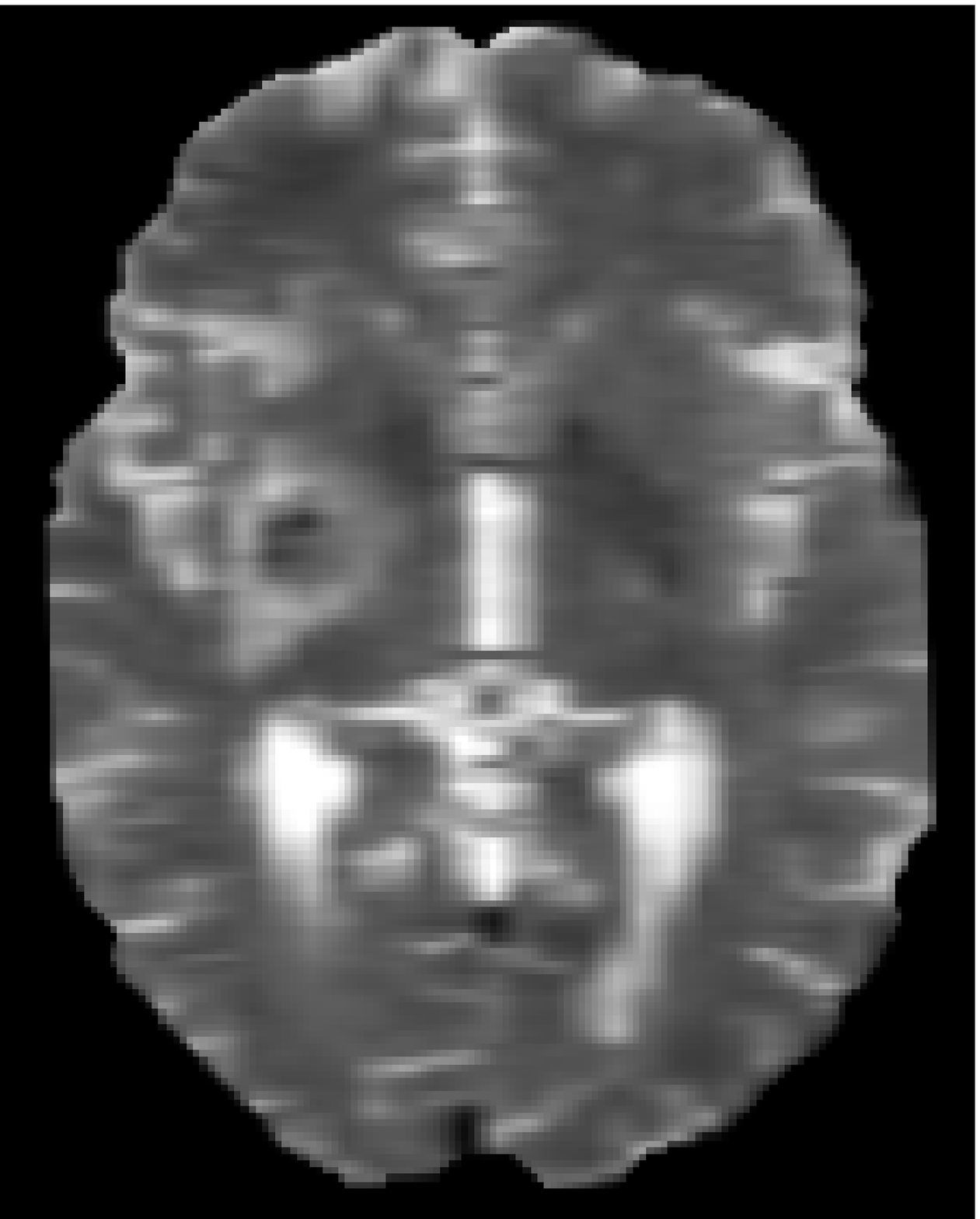}
		\caption{Sym. T2}
	\end{subfigure} 
	\begin{subfigure}[b]{\modsubfig}
		\includegraphics[width=\textwidth,trim={0px 0px 0px 0px},clip]{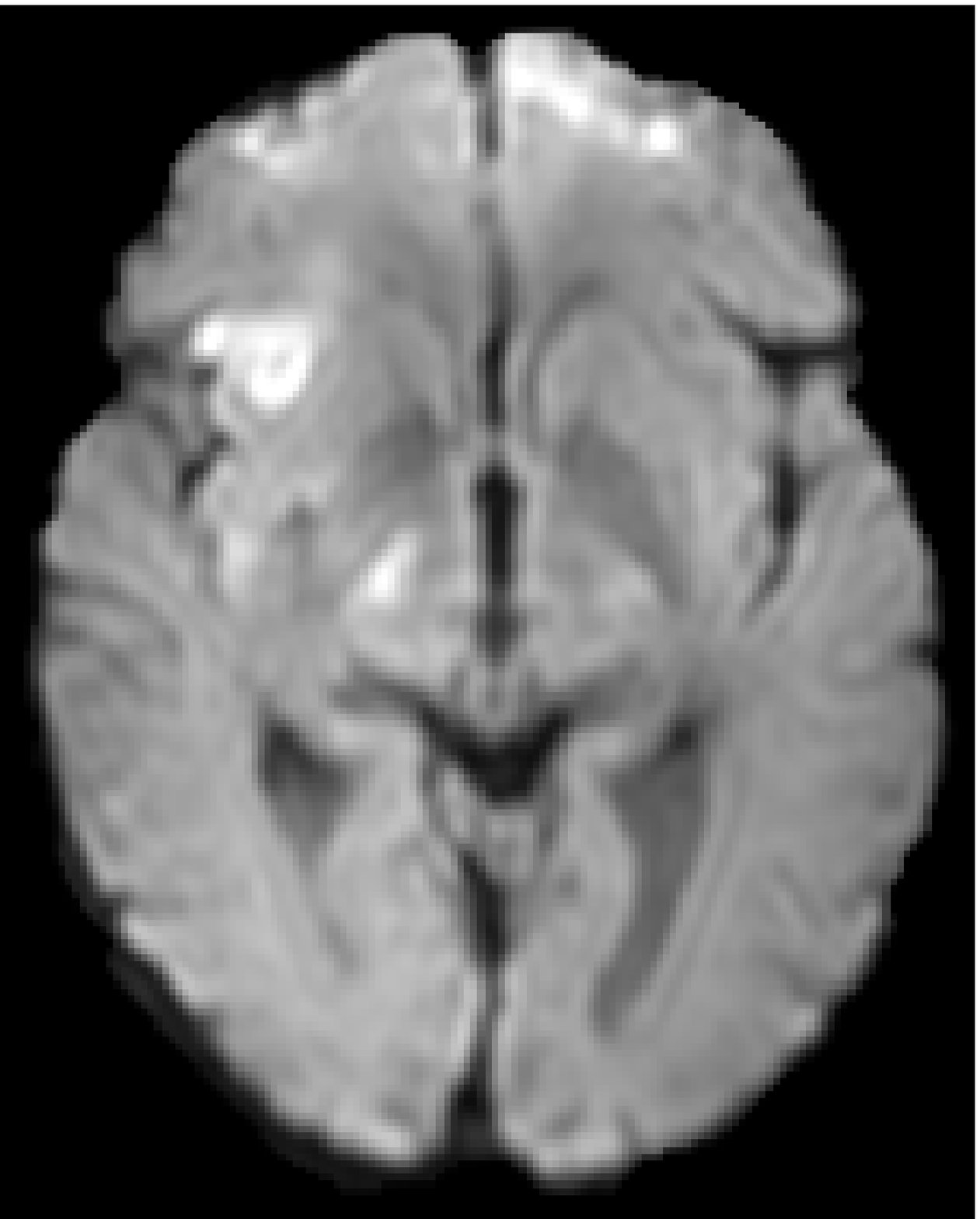}
		\caption{Sym. DWI}
	\end{subfigure} 
	\begin{subfigure}[b]{\modsubfig}
		\includegraphics[width=\textwidth,trim={0px 0px 0px 0px},clip]{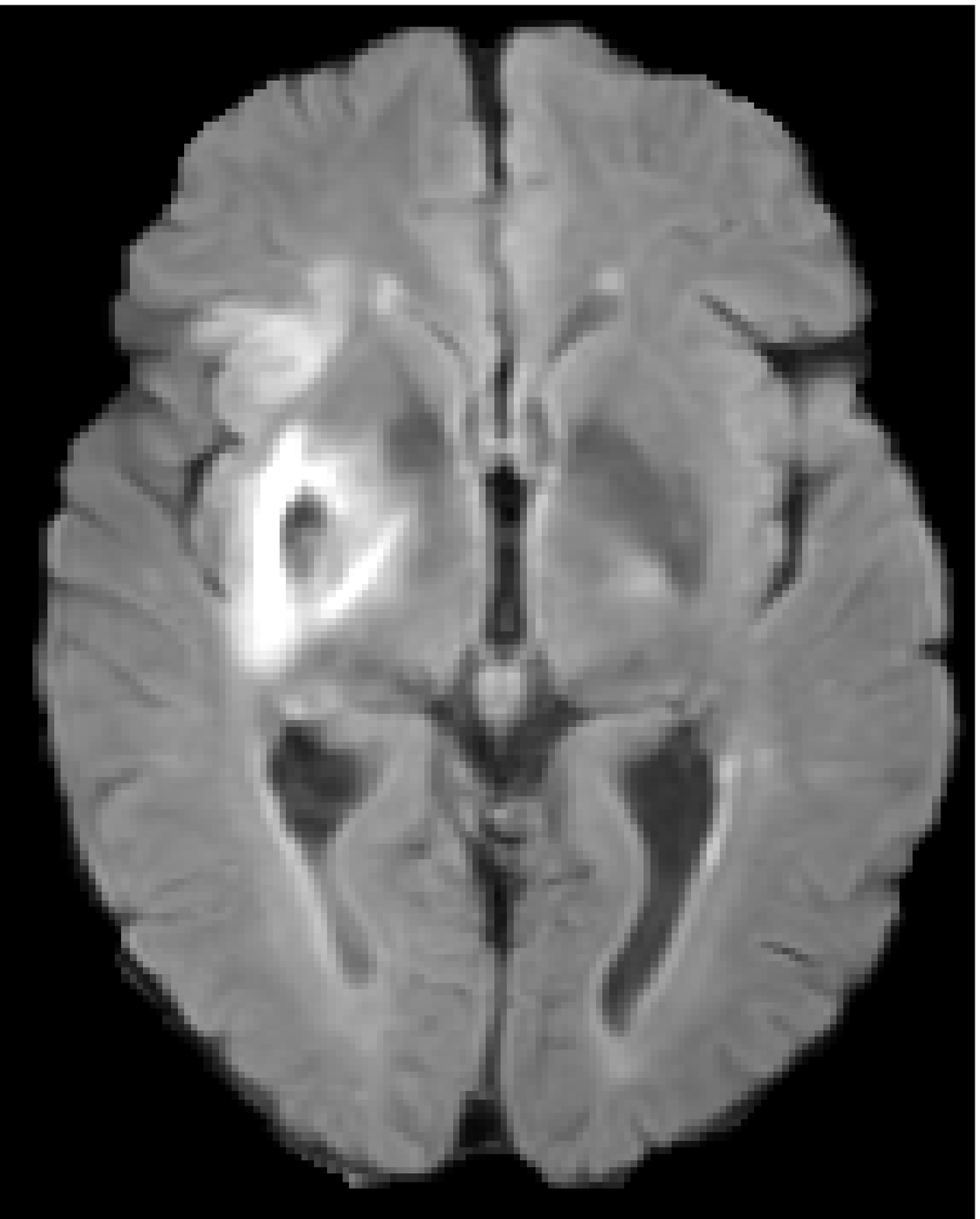}
		\caption{Sym. FLAIR}
	\end{subfigure}
	\caption{Provided and symmetrically augmented modalities from case 2 of the SISS training images.}
	\label{fig:sym}
\end{figure*}

\subsection{CNN architecture}

\begin{figure*}
	\centering
	\includegraphics[width=1.0\textwidth]{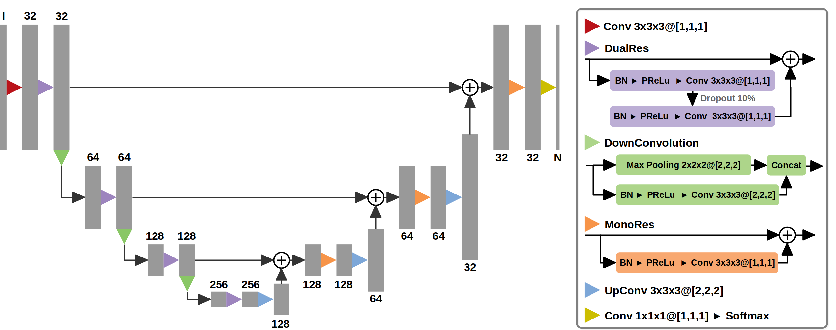}
	\caption[]{Employed U-Net based architecture using 3D convolutions, 4 resolution steps and 32 base filters. The architecture consists of an asymmetrical encoder-decoder network using long and short residual connections. For the convolutional layers,  K\textsubscript{x}$\times$K\textsubscript{y}$\times$K\textsubscript{z}@[S\textsubscript{x},S\textsubscript{y},S\textsubscript{z}] indicates the kernel and stride dimensions in each axis. The number of channels is indicated above or under each feature map. In the input and output feature maps, I and N denote the number of image modalities and segmentation classes respectively.}
	\label{fig:sunet}
\end{figure*}

The used architecture, illustrated in Figure~\ref{fig:sunet}, is a 3D asymmetrical encoder-decoder network based on the U-Net \cite{Ronneberger2015} architecture and its 3D extension, the 3D U-Net \cite{Cicek2016}. Additionally, we also use short and long residual connections as used by the 2D uResNet architecture \cite{Guerrero2018uresnet} for chronic stroke in MRI. The asymmetry comes from the number of parameters found in the encoder and decoder branches, with 75\% and 25\% of the parameters respectively. We use residual blocks with two convolutional layers in the encoder and with a single convolutional for the decoder. It has been shown that the decoder's role is not as critical for segmentation, mainly upsampling the work of the encoder and fine-tuning the details \cite{Paszke2016}. Additionally, instead of the more typical rectified linear unit (ReLU) \cite{NairVinodandHinton2010} we use in our residual blocks a parametric version, the PReLU non-linearity \cite{He2015}, as suggested by Paszke et al \cite{Paszke2016}. We perform downsampling in each resolution step by concatenating the result of a max pooling operation and strided convolution as proposed by Szegedy et al \cite{Szegedy2015}. This strategy avoids representational bottlenecks while keeping the number of parameters contained. Finally, upsampling in the decoder branch is performed with the use of transposed convolutions. 

\subsection{Class imbalance handling}
\label{subsec:imbalance}
The class imbalance issue is caused by the typically smaller extent of the lesion class as compared with the rest of healthy tissue class. If no deliberate action is taken, the training set will be composed mostly from examples of healthy tissue and few from the lesion. This would induce a biased learning that would harm the segmentation performance. To alleviate this issue, we use a combination of small patches with a balanced training patch sampling and a difficulty weighted loss function. The employed loss function, the Focal loss \cite{Lin2018}, is a dynamically weighted extension of the cross entropy loss defined as: 

\begin{equation}\label{eq:score}
\text{FL}(p_\text{t}) = - \alpha_\text{t} (1-p_\text{t})^\gamma \log(p_\text{t})
\end{equation}

\noindent where $p_\text{t}$ and $\alpha_\text{t}$ are the predicted probability and weight for class t respectively. This function is dynamically weighted inversely proportional to the prediction confidence, so the network learns less from confident classifications and more from misclassified examples. In this way, class imbalance is alleviated as the network stops learning from the larger amount of healthy examples while still learning from the less common lesioned tissue. We use the Focal loss default parameters as suggested by Lin et al \cite{Lin2018}, with scaling factor $\gamma=2$ and class weights $\alpha_0=0.25$ and $\alpha_1=0.75$ for the healthy and lesion classes respectively.

\begin{figure*}[t]
	\centering
	\includegraphics[width=1.0\textwidth]{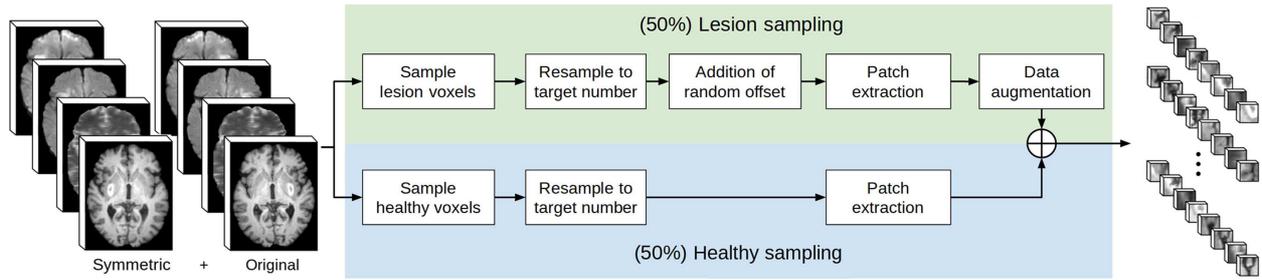}
	\caption[]{Diagram of the used training patch sampling strategy for acute stroke related tasks that considers the anatomy and pathophysiology of stroke lesions. The extracted patches are of size $24\times24\times16$ and include all input modalities.}
	\label{fig:samplingdiagram}
\end{figure*}

The use of patches allows using a training sampling strategy that can undersample the healthy class and oversample the lesion for a more balanced class representation. The employed strategy is an extension of two recently proposed ones for brain lesions \cite{Kamnitsas2017dm} and chronic stroke \cite{Guerrero2018uresnet}. In practice, a goal number of patches to extract is set per patient, as we aim to have a balanced patch representation of each case. Then, 50\% of the training patches are extracted centered on voxels corresponding to healthy tissue and the other 50\% on lesion. These are sampled at regular spatial steps to ensure that all parts of the brain are equally represented. The voxels sampled from the lesion class have a random offset added to increase representation of the region surrounding the lesion, the benign oligemia. As suggested by Guerrero et al \cite{Guerrero2018uresnet}, the offset is limited to half of the patch size to ensure the originally sampled voxel remains in the final extracted patch. For patients with smaller lesions, a combination of several patch extractions from the same lesion voxel and data augmentation is done to ensure the number is reached. The same sampled voxel will actually produce different patches since lesion voxels have a random offset applied. Finally, patches are extracted centered on these voxels. Additionally, for the lesion sampled patches, data augmentation is applied with five anatomically feasible operations including sagittal reflections and 90\textdegree, 180\textdegree~and 270\textdegree~axial rotations. A diagram summarizing the described strategy is depicted in Figure~\ref{fig:samplingdiagram}.

Despite the balancing effects of the Focal loss and training patch sampling, the segmentation performance is still reduced when bigger patch sizes are considered. Since there are much fewer lesion voxels than healthy ones, bigger patches tend to include more healthy class voxels and further worsen class imbalance in the training set. The employed patch size of $24\times24\times16$, determined empirically, offers the best compromise between receptive field and worsened imbalance for the considered datasets.

\subsection{Network training}

For training the randomly initialized network weights, we first extract patches to build the training and validation sets. As stated in Section~\ref{subsec:imbalance}, we use patches of size $24\times24\times16$ sampled with a balanced patch sampling strategy. During training, we use the Focal loss \cite{Lin2018} along with the Adadelta optimizer \cite{Zeiler2012}, to avoid costly grid search of a learning rate, with a batch size of 16 patches. This optimizer requires no manual tuning of parameters and appears robust to noisy gradient information, different model architecture choices, various data modalities and selection of hyper-parameters. Moreover, to prevent overfitting we use the early stopping technique by monitoring the performance on a validation set at the end of each epoch. In this way, the training is interrupted when the monitored metric reaches a local minimum, which means no more generalizable knowledge is being learned from the training images. The sum of the L1 loss and error rate on the validation set is used as the monitored metric with a patience of 8 epochs.

\subsection{Segmentation and post-processing}
\label{subsec:segmentation}
Once the network weights have been trained, to segment a new volume patches are first extracted from every part of the image and forward passed through the network. These are sampled uniformly with a regular extraction step of $4 \times 4 \times 1$ so that all parts of the brain are predicted. The resulting patch probabilities are then combined in a common space  preserving their original spatial location to produce the whole volume probability map. In our case, the combination is performed per voxel by averaging the class probabilities of the various patches. Furthermore, some degree of overlap between the extracted patches is used since the extraction step is smaller than the patch size. Therefore, the same voxel is labeled seen in different neighborhoods and the resulting class probabilities are averaged. This technique reduces the need for post-processing steps as it provides coherently spatial labels without block artifacts.


Finally, the probability maps are binarized by thresholding the lesion class probabilities and then performing a connected component filtering by lesion volume. The variable threshold $T_h$ can compensate over/under confident networks while the minimum lesion size $S_{min}$, measured in number of voxels, takes advantage of lesion priors to minimise false positives. In practice, the probability maps are binarised using the same threshold and minimum lesion size for each evaluation. These are found through grid search after all networks have been trained to offer the best compromise between the desired evaluation metrics.

\subsection{Implementation details}

The proposed method has been implemented with Python, using the Torch scientific computing framework \cite{Paszke2017}. All experiments have been run on a GNU/Linux machine running Ubuntu 18.04 with 64GB of RAM memory and an Intel\textsuperscript{®} Core\textsuperscript{TM} i7-7800X CPU. The network training and testing has been done with an NVIDIA TITAN X GPU (NVIDIA corp, United States) with 12GB G5X memory. 

\section{Evaluation and results}

We perform a quantitative and qualitative evaluation with both a cross-validation experiment and a blind external evaluation using the challenge web platform. The metrics used in the quantitative evaluations will be the ones provided by the online platform. These include the Dice similarity coefficient (DSC) \cite{Dice1945}, sensitivity, positive predictive value (PPV) and Hausdorff distance (HD). The DSC measures the relative overlap of the segmentation with the ground truth and is used as a measure of segmentation performance. The sensitivity and PPV measure different properties relative to the lesion class segmentation. On the one hand, the sensitivity  evaluates the percentage of gold standard lesion correctly labeled as such. On the other hand, the PPV measures the fraction of lesion class predictions that are correct. Finally, the Hausdorff distance can be intuitively seen as a measure of the \textit{largest} border error between the segmentation and ground truth.

\subsection{Cross-validation experiment with training images}

The purpose of the cross-validation experiment is to quantitatively asses the main introduced improvements of the proposed methodology against a Baseline approach without them. For the Baseline approach, we use the proposed methodology without the class imbalance handling nor the data pre-processing step. Instead we use the crossentropy loss and training patch sampling as described in \cite{Kamnitsas2017dm}, using $24\times24\times16$ patches without any addition of a random offset. We then evaluate the effects of a Balanced approach that only uses the class imbalance handling described in Section~\ref{subsec:imbalance}, without performing symmetric modality augmentation. Finally, the Proposed approach also adds the data pre-processing step to implement the complete proposed methodology. 

Each evaluation is performed in 4 folds, adjusting the number of cases per fold accordingly, with the same training procedure for both the SISS and SPES datasets. To build the patch training set for each fold, 10~000 patches per case are extracted from the training images summing approximately 260~000 patches in total. Once the networks from each fold have been trained and the probability maps generated for all training images, the post-processing parameters $T_h$ and $S_{min}$ are found through grid search to optimize the desired metrics across all folds. We consider the range of thresholds $T_h$ from 0.1 to 0.9 and minimum lesion size $S_{min}$ from 10 to 1000 voxels. More specifically, we choose the parameter combination that jointly maximizes the average DSC and HD, the two metrics used to determine the 2015 ISLES workshop results. In practice, a combined score is computed as:

\begin{equation}
	\text{Score} = \dfrac{ \text{DSC} * \left( 1 - \dfrac{\text{HD}}{\text{HD}_{\text{max}}} \right) }{\text{DSC} + \left( 1 - \dfrac{\text{HD}}{\text{HD}_{\text{max}}} \right)}
\end{equation}

\noindent where $\text{HD}_{\text{max}}$, set to 200 voxels, is used to normalize the HD metric to the range between 0 and 1.

\subsubsection{SISS sub-task results}

The evaluation metrics of the cross-validation experiment using the SISS dataset can be found in Table~\ref{table:crossval}. With respect to the Baseline, the Balanced approach significantly improves the Hausdorff distance (p $<$ 0.01) with marginal improvements in other metrics. When the symmetrically augmented modalities are further considered, the Proposed approach achieves significantly better DSC, PPV and HD (p $<$ 0.02) as compared with the Baseline. However, despite the improvement in evaluation metrics, the Proposed approach needs a more restrictive minimum lesion size of 200 voxels to maximize the score as compared with the Baseline, which only filtered lesions smaller than 50 voxels.

\begin{table*}[t]
	\centering
	\caption{Cross-validation experiment evaluation metrics on the SISS and SPES sub-tasks. The post-processing parameters $T_h$ and $S_{min}$ are found through grid search to maximize the score defined in Equation~\ref{eq:score}.}
	\begin{tabular}{lrrrrrr}
		\hline
		Approach & $T_h$ & $S_{min}$ & DSC & PPV & Sensitivity & HD \\
		\hline	
		SISS sub-task & \multicolumn{6}{l}{~} \\
		\hline
		Baseline & 0.4 & 50 & 0.64 ± 0.22 & 0.69 ± 0.27 & 0.68 ± 0.21 & 43.7 ± 32.6 \\
		Balanced & 0.4 & 200 & 0.67 ± 0.21 & 0.73 ± 0.22 & 0.69 ± 0.23 & 30.9 ± 28.9 \\
		Proposed & 0.5 & 200 & 0.71 ± 0.19 & 0.78 ± 0.20 & 0.67 ± 0.22 & 29.5 ± 29.5 \\
		\hline
		SPES sub-task & \multicolumn{6}{l}{~} \\
		\hline
		Baseline & 0.6 & 500 & 0.80 ± 0.17 & 0.82 ± 0.21 & 0.82 ± 0.19 & 11.1 ± 6.9 \\
		Balanced & 0.4 & 500 & 0.82 ± 0.15 & 0.84 ± 0.14 & 0.85 ± 0.17 & 12.4 ± 7.6 \\
		Proposed & 0.5 & 200 & 0.82 ± 0.16 & 0.85 ± 0.13 & 0.85 ± 0.17 & 11.2 ± 7.3 \\
		\hline
	\end{tabular}
	\label{table:crossval}
\end{table*}

%

\begin{figure*}[t]
	\centering
	\begin{subfigure}[b]{0.228\textwidth}
		\includegraphics[width=\textwidth,trim={0px 0px 0px 0px},clip]{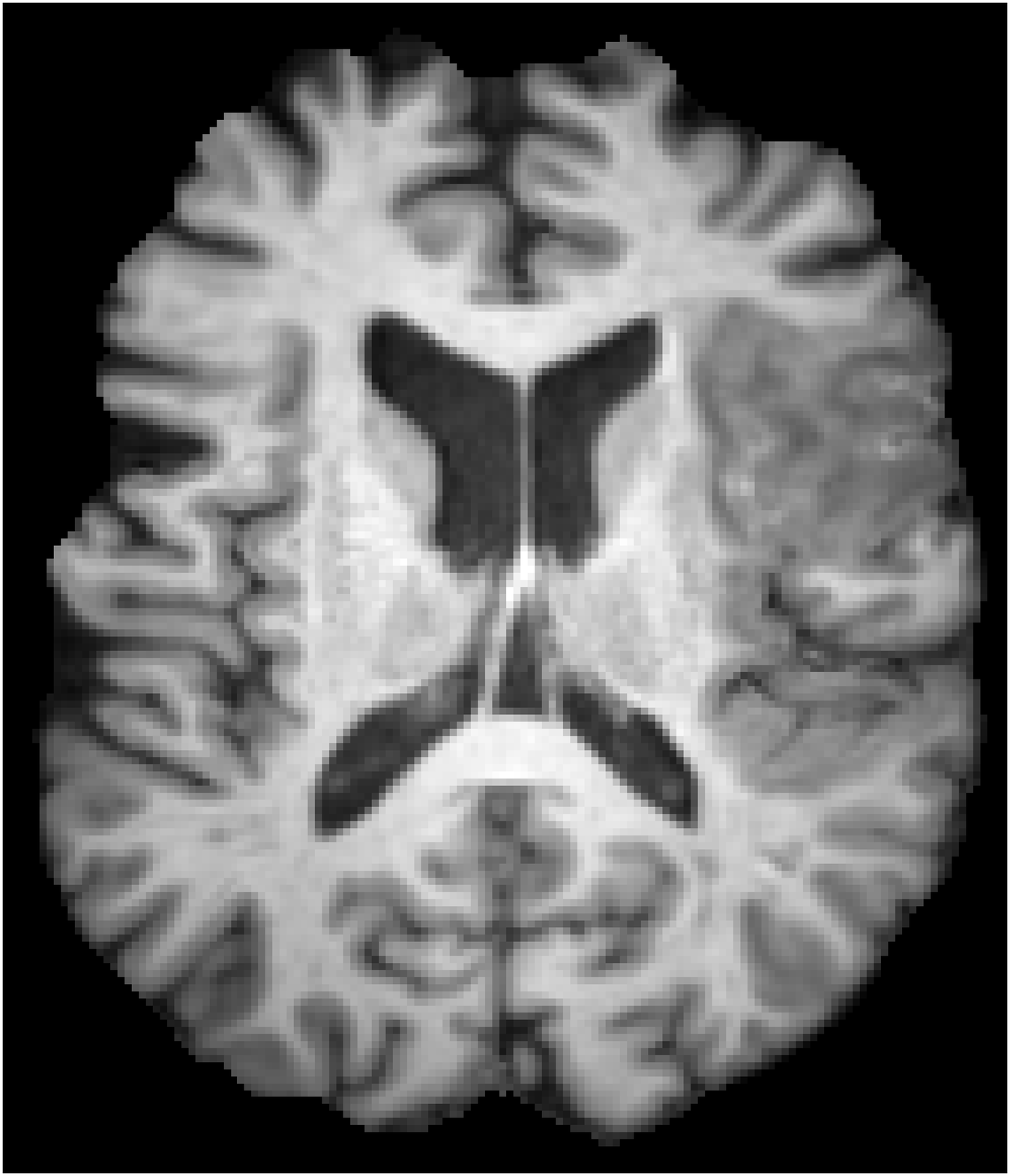}
	\end{subfigure} 
	\quad
	\begin{subfigure}[b]{0.2\textwidth}
		\includegraphics[width=\textwidth,trim={0px 0px 0px 0px},clip]{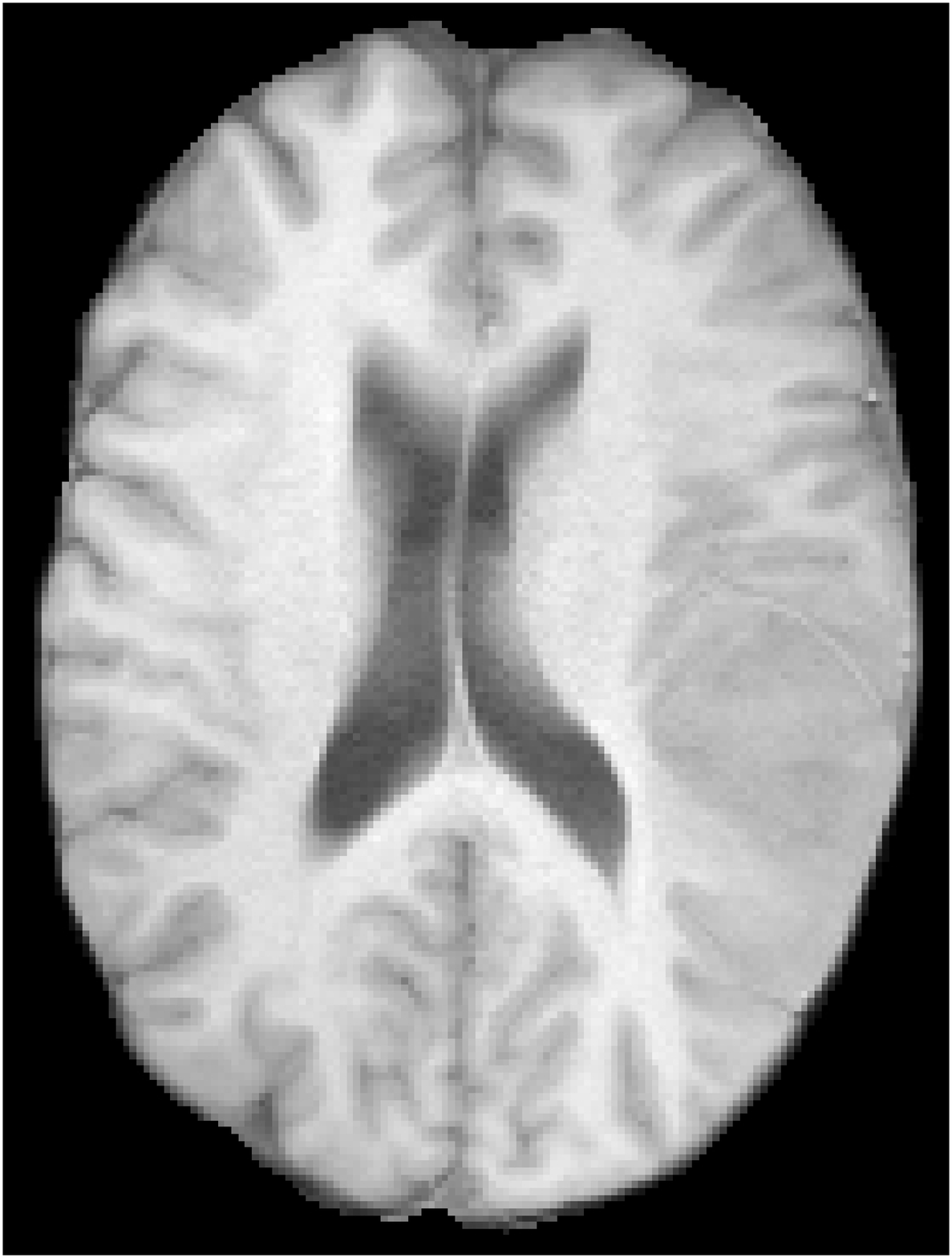}
	\end{subfigure} 
	\quad 
	\begin{subfigure}[b]{0.227\textwidth}
		\includegraphics[width=\textwidth,trim={0px 0px 0px 0px},clip]{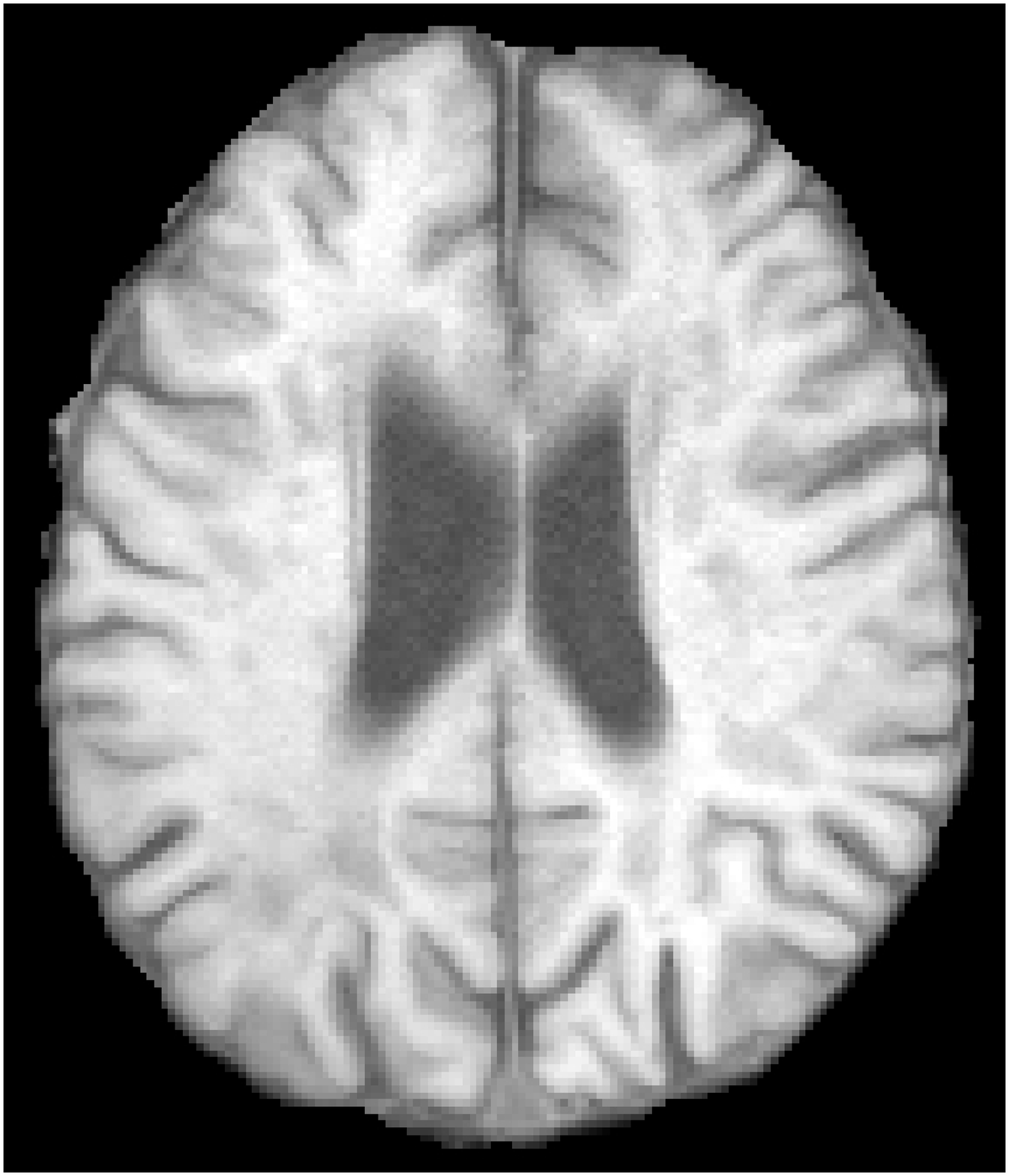}
	\end{subfigure}  
	\quad
	\begin{subfigure}[b]{0.211\textwidth}
		\includegraphics[width=\textwidth,trim={0px 0px 0px 0px},clip]{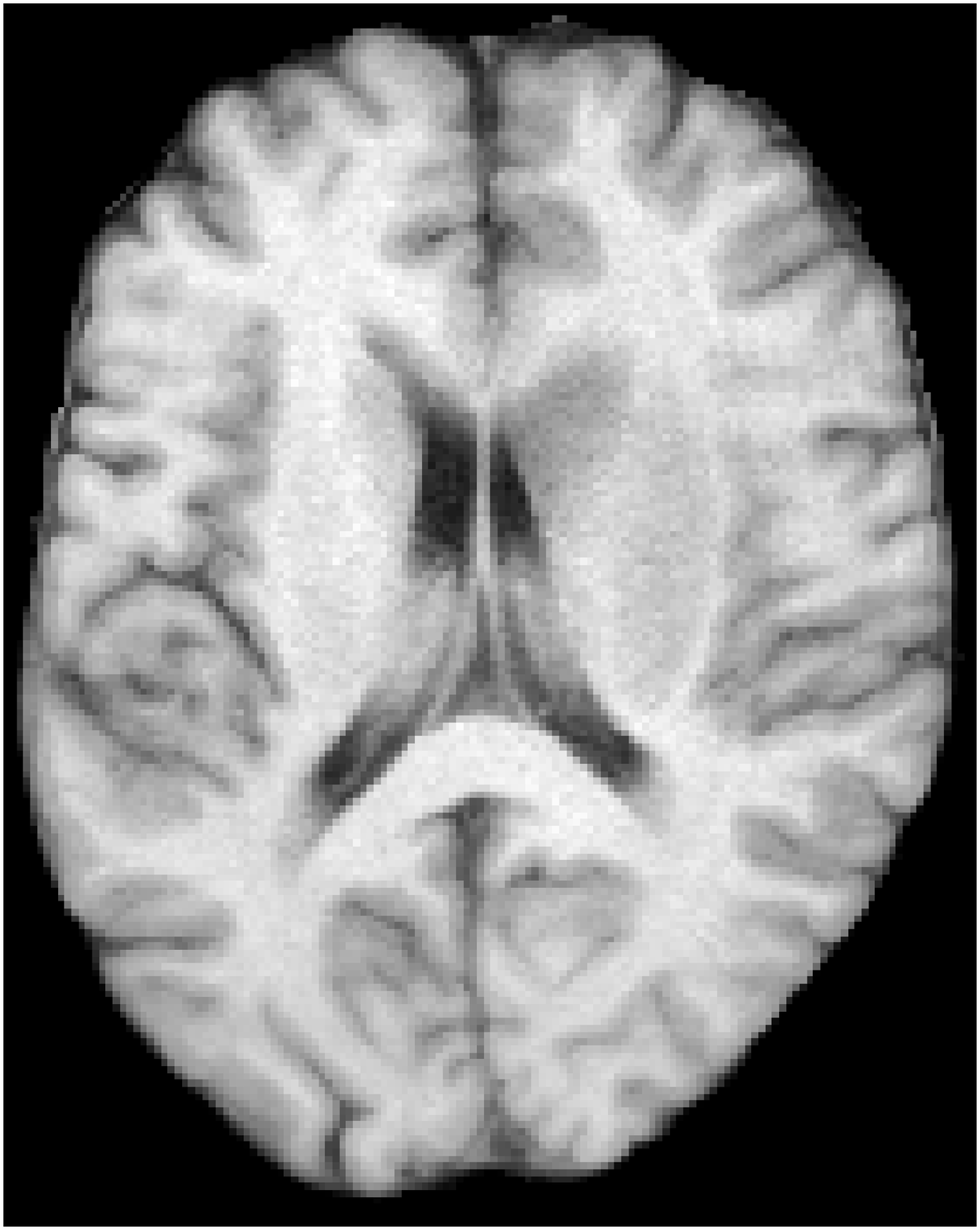}
	\end{subfigure}
	\\[0.5em]
	\begin{subfigure}[b]{0.228\textwidth}
		\includegraphics[width=\textwidth,trim={0px 0px 0px 0px},clip]{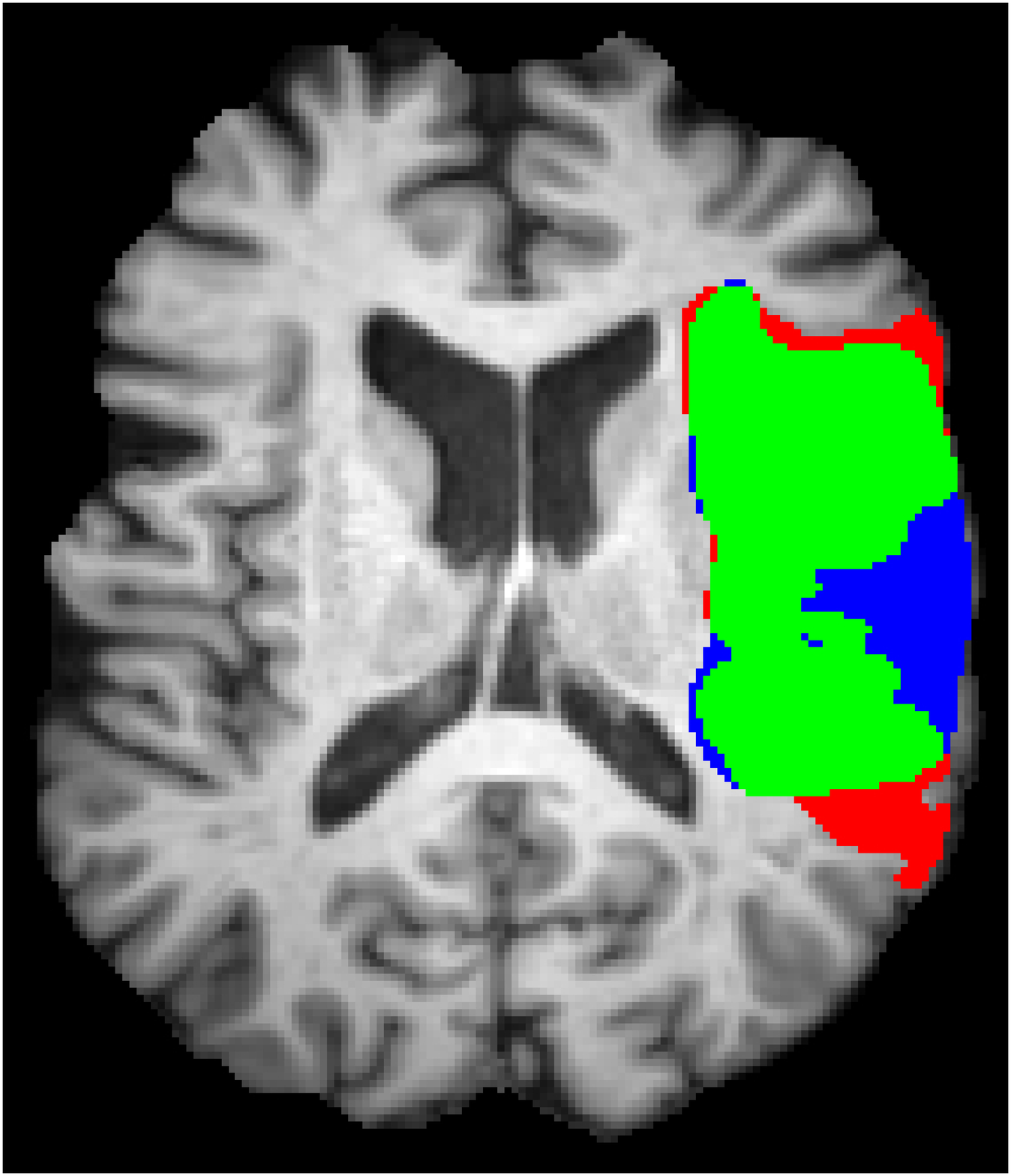}
		\caption{Case 5}
	\end{subfigure}
	\quad
	\begin{subfigure}[b]{0.2\textwidth}
		\includegraphics[width=\textwidth,trim={0px 0px 0px 0px},clip]{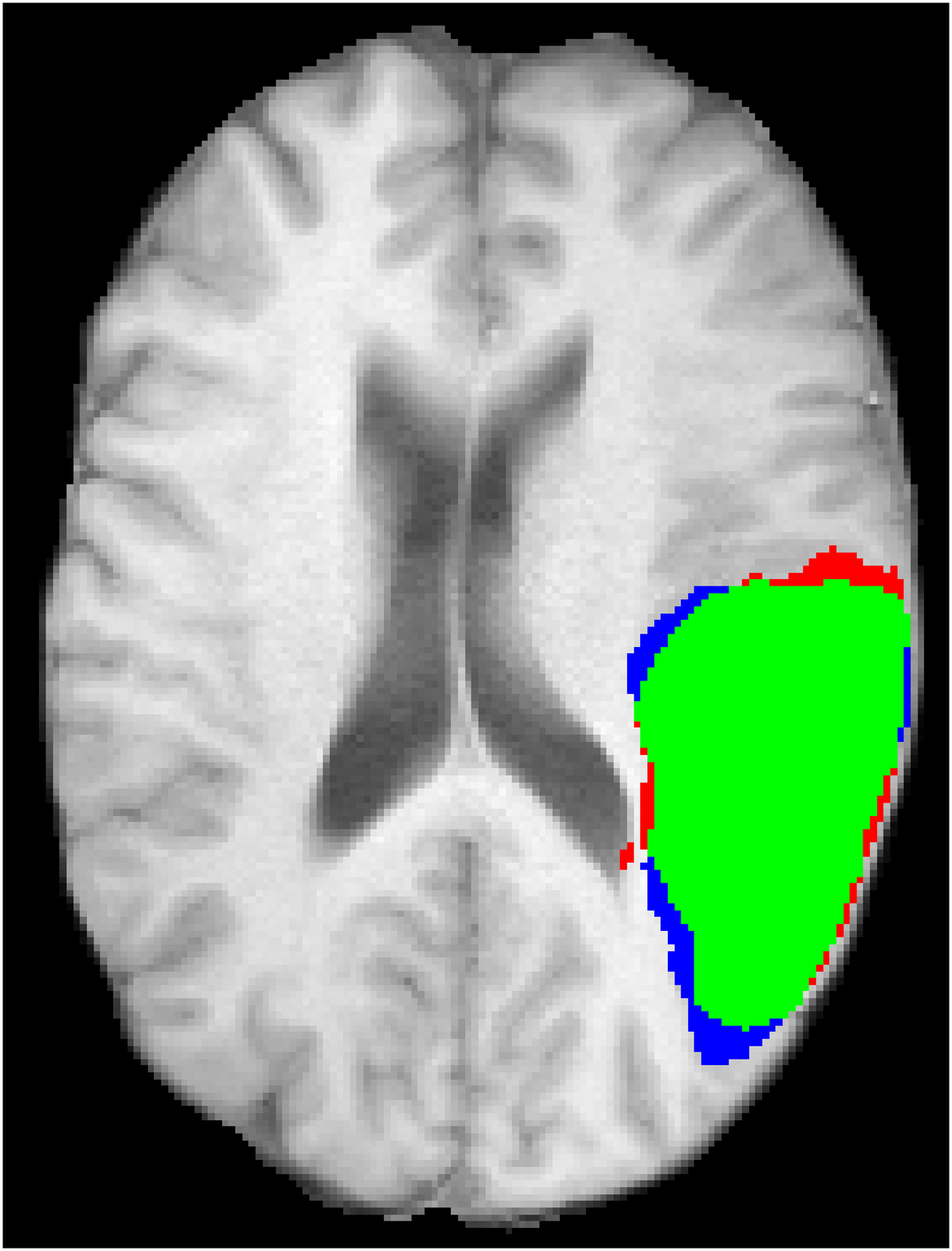}
		\caption{Case 9}
	\end{subfigure} 
	\quad
	\begin{subfigure}[b]{0.227\textwidth}
		\includegraphics[width=\textwidth,trim={0px 0px 0px 0px},clip]{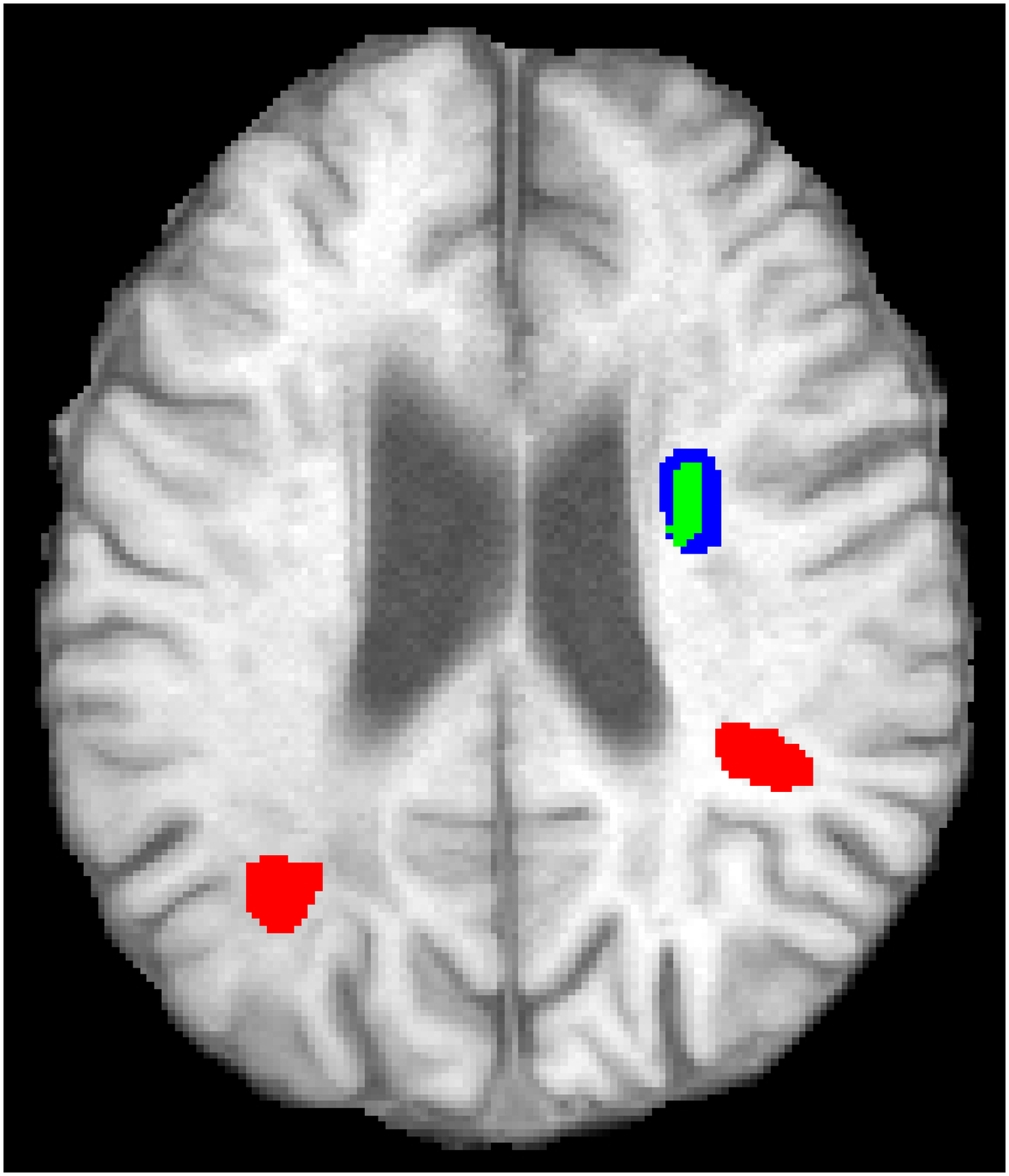}
		\caption{Case 13}
	\end{subfigure}
 	\quad
	\begin{subfigure}[b]{0.211\textwidth}
		\includegraphics[width=\textwidth,trim={0px 0px 0px 0px},clip]{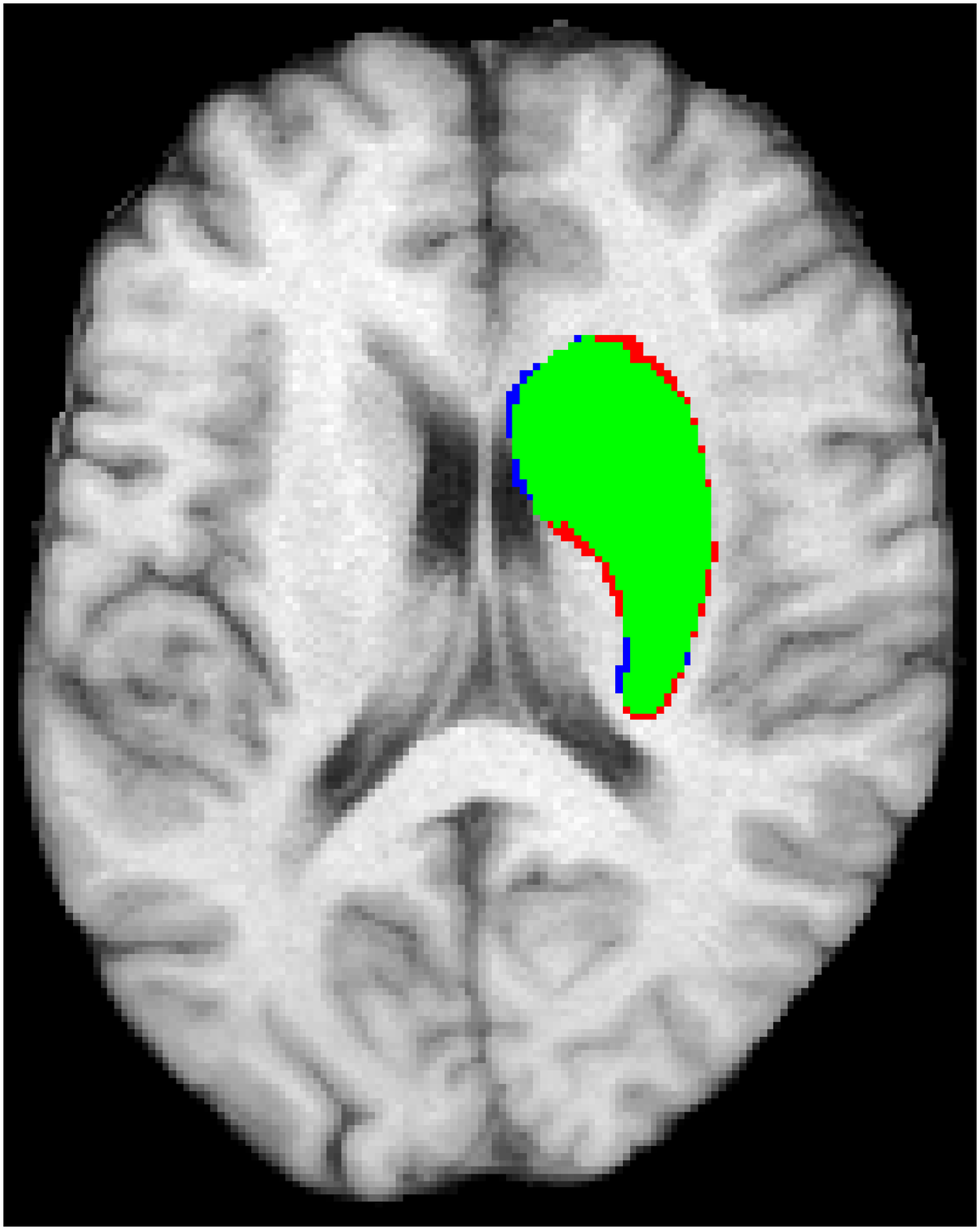}
		\caption{Case 15}
	\end{subfigure}
	\caption{Output segmentation masks of representative cases from the training images of ISLES 2015 SISS dataset. On all images, true positives are denoted in green, false positives in red and false negatives in blue.}
	\label{fig:qsiss}
\end{figure*}

Representative examples of the qualitative results from the proposed method can be found in Figure~\ref{fig:qsiss}. Cases 9 and 15 represent the overall results of the proposed methodology, correctly detecting the lesions in most cases with an outline that approximates the provided gold standard. Among the observed limitations are inaccurate borders and over/under segmentation of certain regions. For instance, in case 5 the lesion was undersegmented due to a heterogeneous appearance of the gold standard lesion while in case 13 two false positive lesions are detected due to the previous existence of chronic stroke lesions with a similar appearance.

\subsubsection{SPES sub-task results}

The evaluation metrics of the cross-validation experiment using the SPES dataset can be found in Table~\ref{table:crossval}. The class imbalance handling used in the Balanced approach significantly improves the sensitivity (p $<$ 0.01) while providing marginal increase on the rest except the Hausdorff distance. When both improvements are simultaneously considered in the Proposed approach, it achieves a significantly better DSC and sensitivity (p $<$ 0.01) than the Baseline. Additionally, the augmented modalities reduce the minimum lesion size $S_{min}$ from 500 to a less restrictive 200 voxels.

\begin{figure*}[t]
	\centering
	\begin{subfigure}[b]{0.232\textwidth}
		\includegraphics[width=\textwidth,trim={0px 0px 0px 0px},clip]{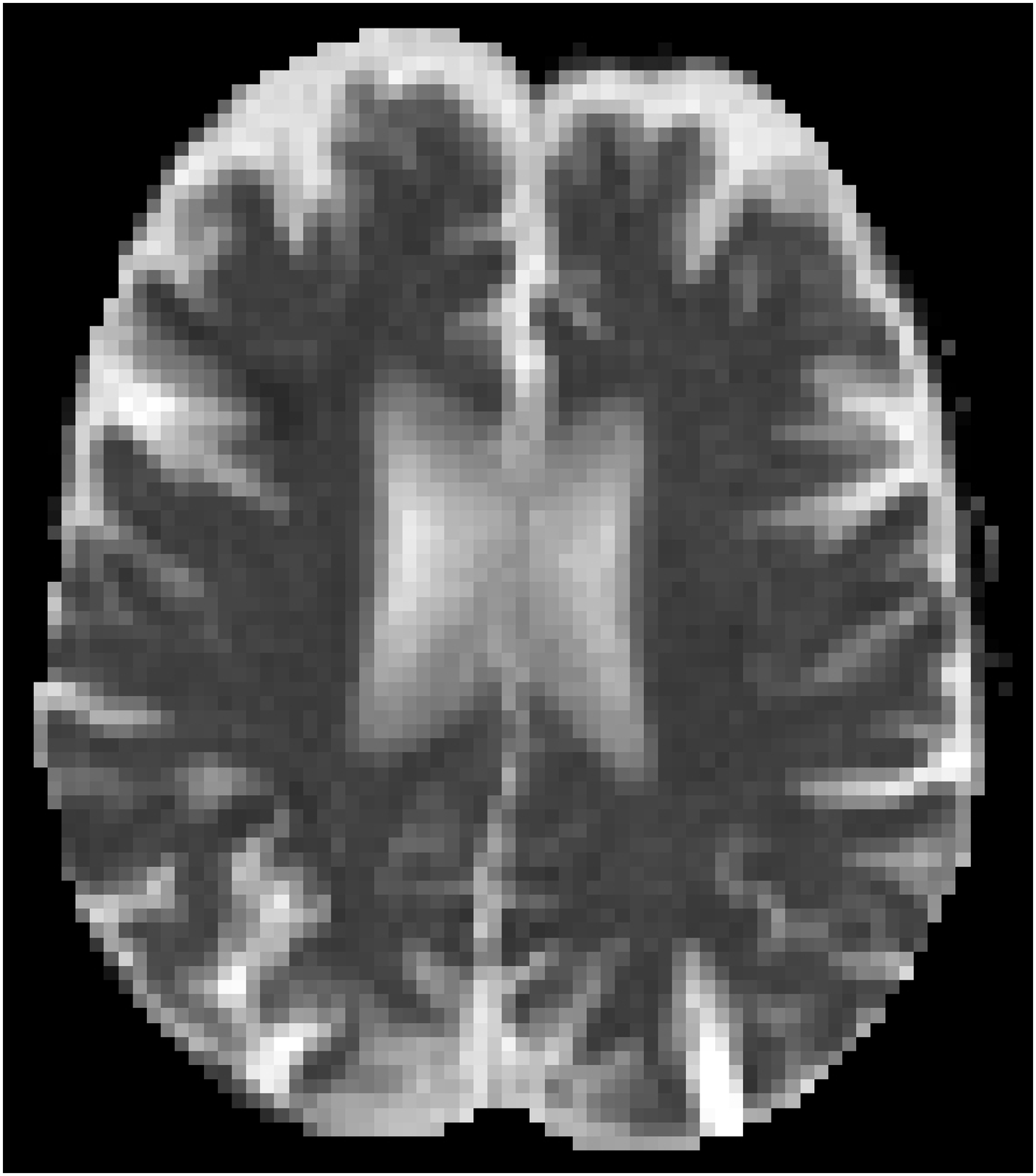}
	\end{subfigure}  
	\quad
	\begin{subfigure}[b]{0.207\textwidth}
		\includegraphics[width=\textwidth,trim={0px 0px 0px 0px},clip]{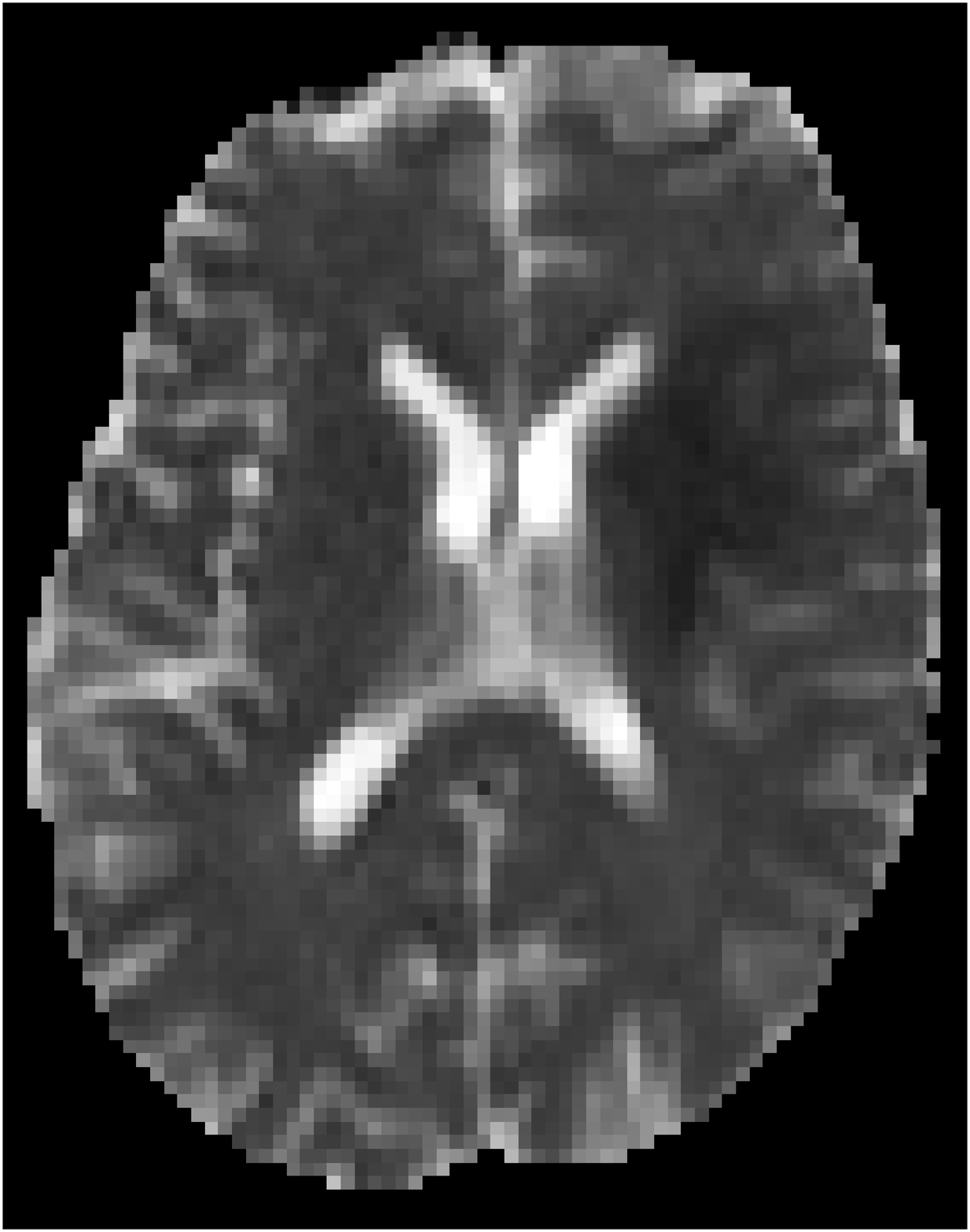}
	\end{subfigure}  
	\quad
	\begin{subfigure}[b]{0.215\textwidth}
		\includegraphics[width=\textwidth,trim={0px 0px 0px 0px},clip]{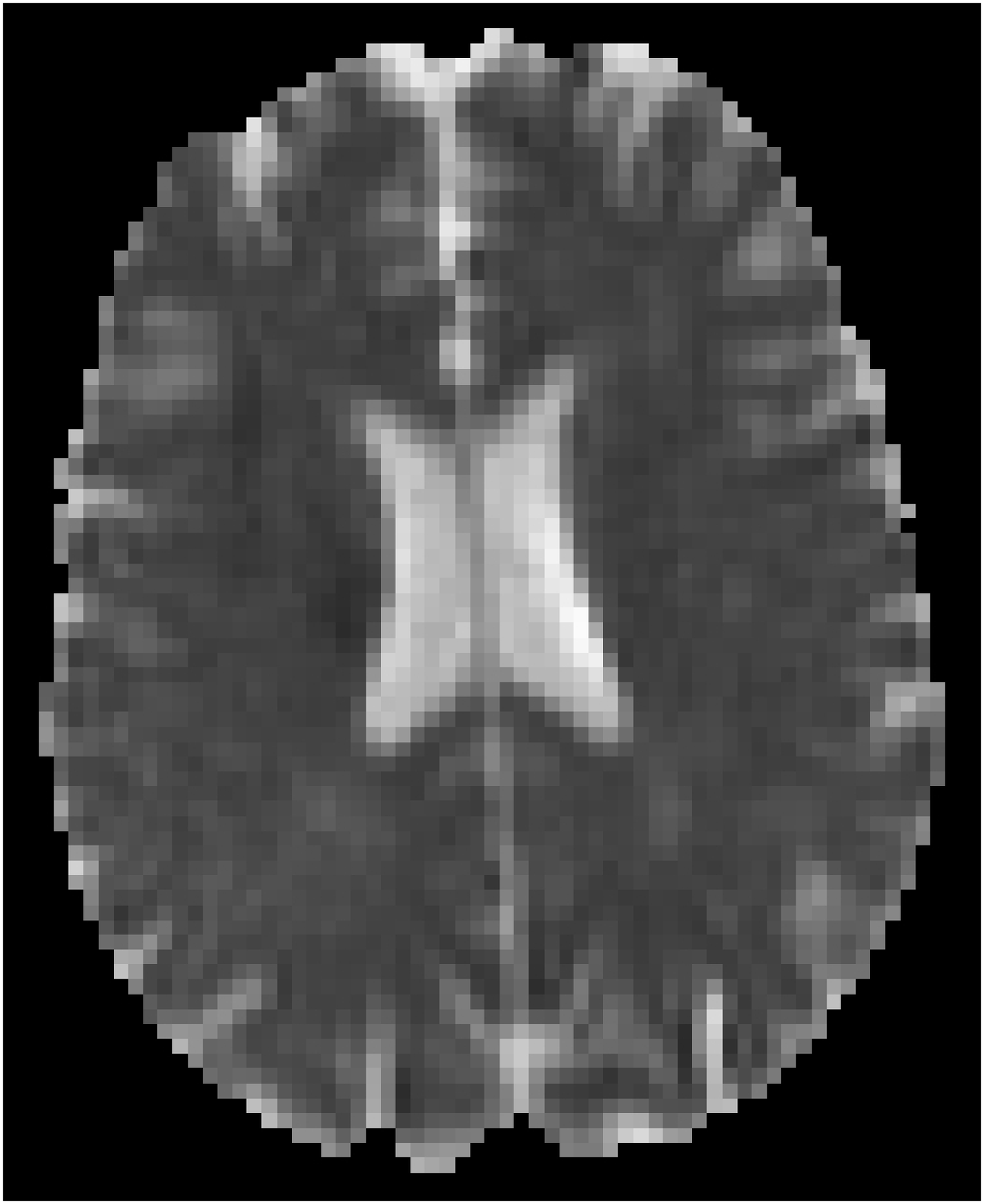}
	\end{subfigure} 
	\quad
	\begin{subfigure}[b]{0.2\textwidth}
		\includegraphics[width=\textwidth,trim={0px 0px 0px 0px},clip]{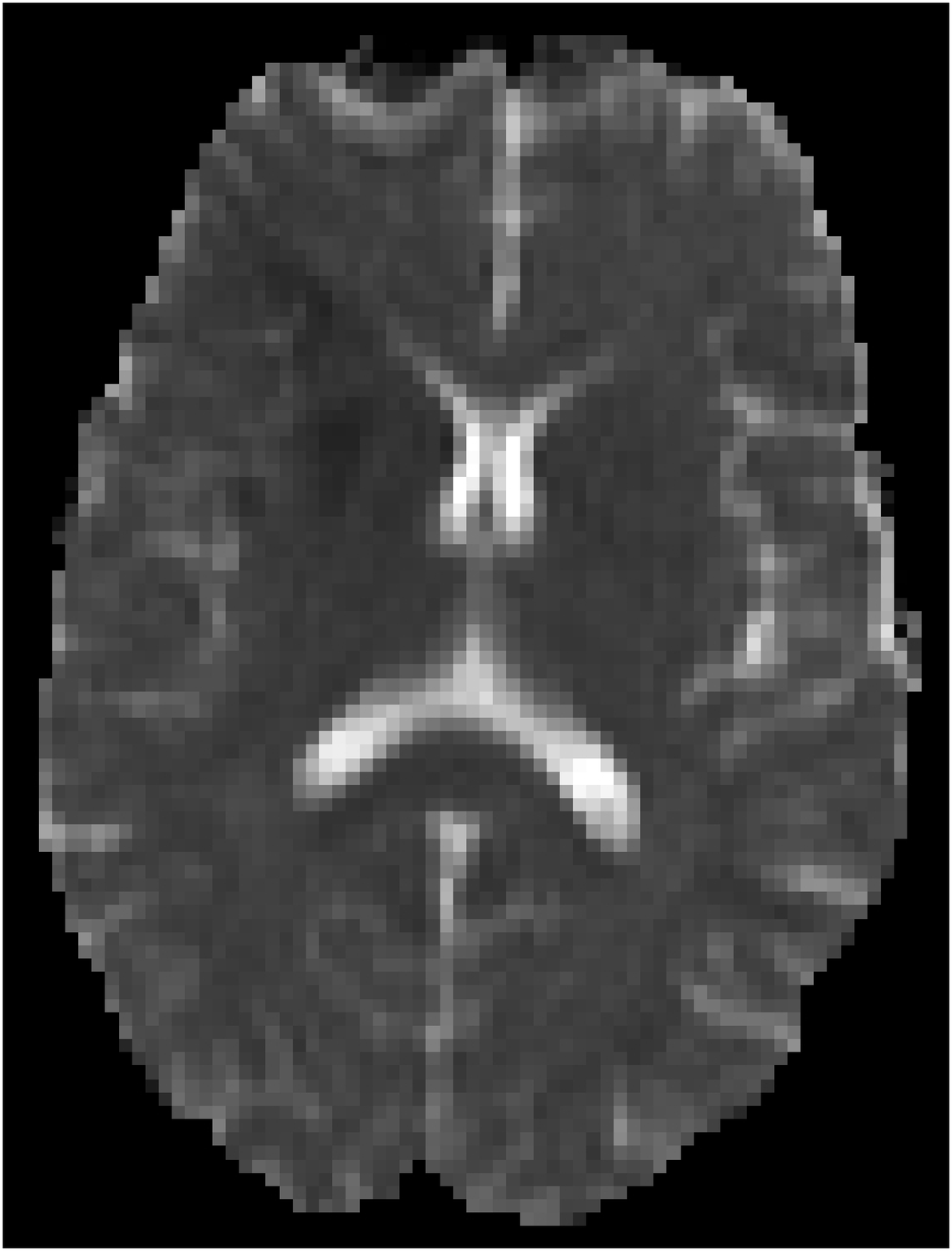}
	\end{subfigure}
	\\ [0.5em]
	\begin{subfigure}[b]{0.232\textwidth}
		\includegraphics[width=\textwidth,trim={0px 0px 0px 0px},clip]{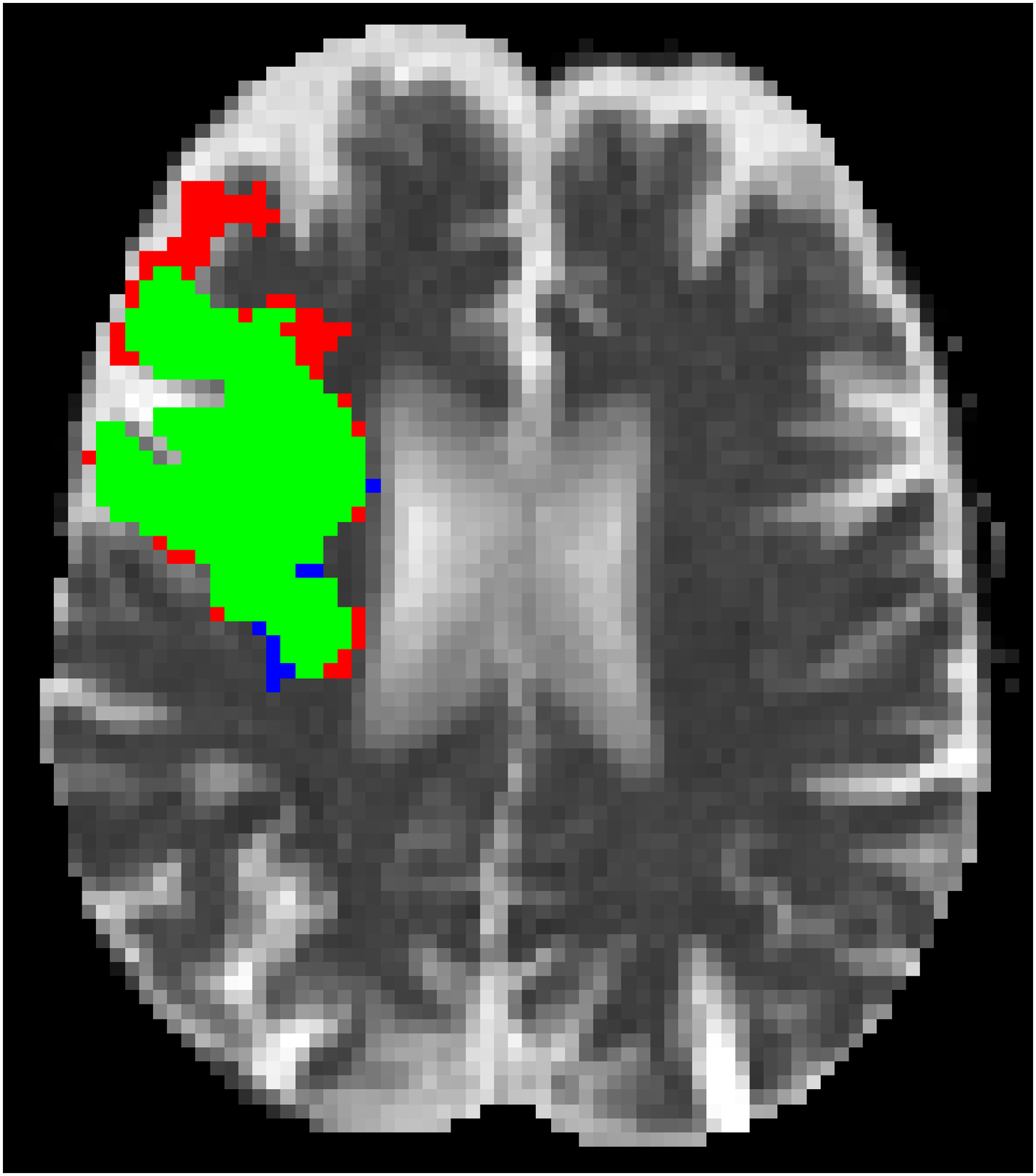}
		\caption{Case 2}
	\end{subfigure} 
	\quad
	\begin{subfigure}[b]{0.207\textwidth}
		\includegraphics[width=\textwidth,trim={0px 0px 0px 0px},clip]{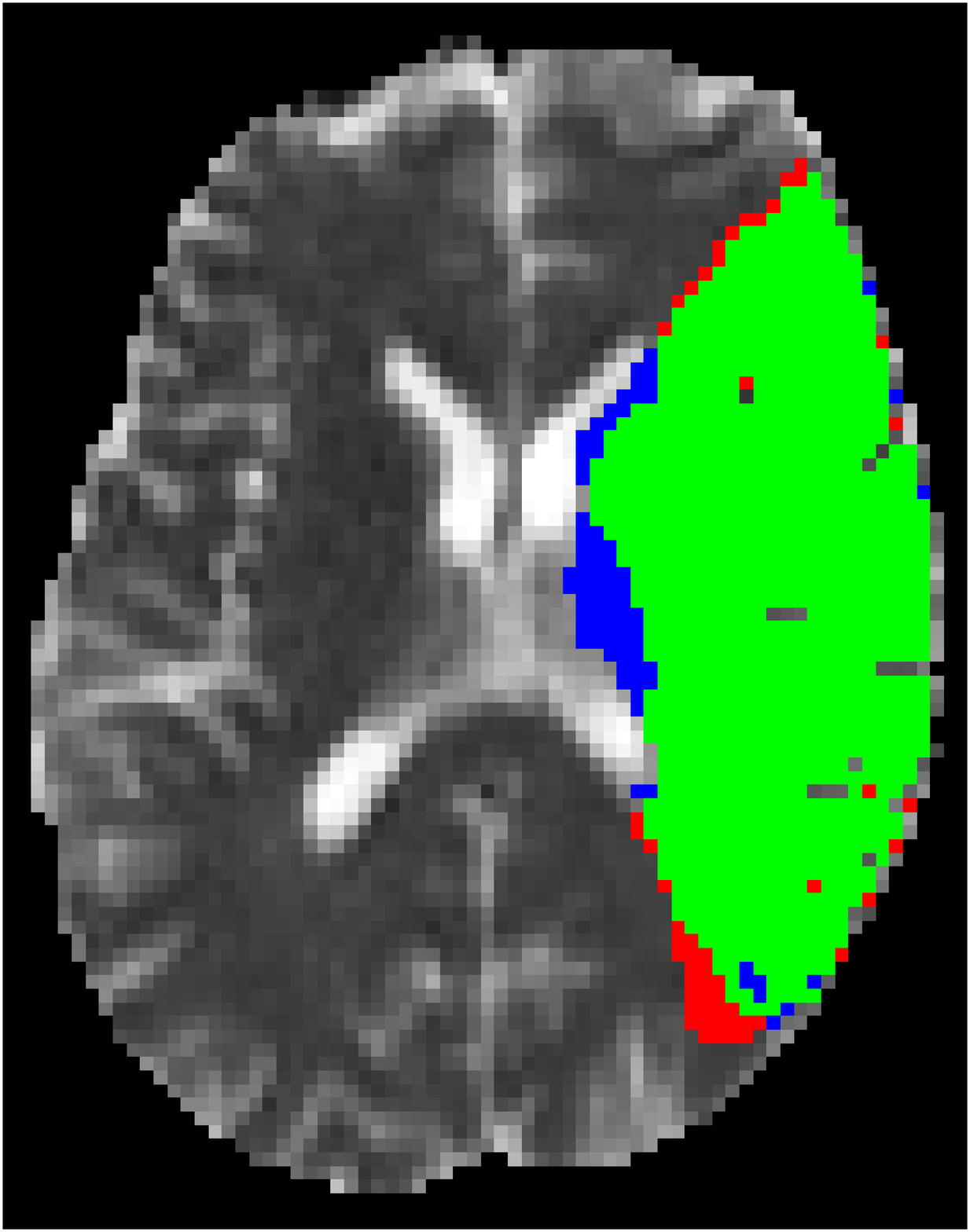}
		\caption{Case 11}
	\end{subfigure} 
	\quad
	\begin{subfigure}[b]{0.215\textwidth}
		\includegraphics[width=\textwidth,trim={0px 0px 0px 0px},clip]{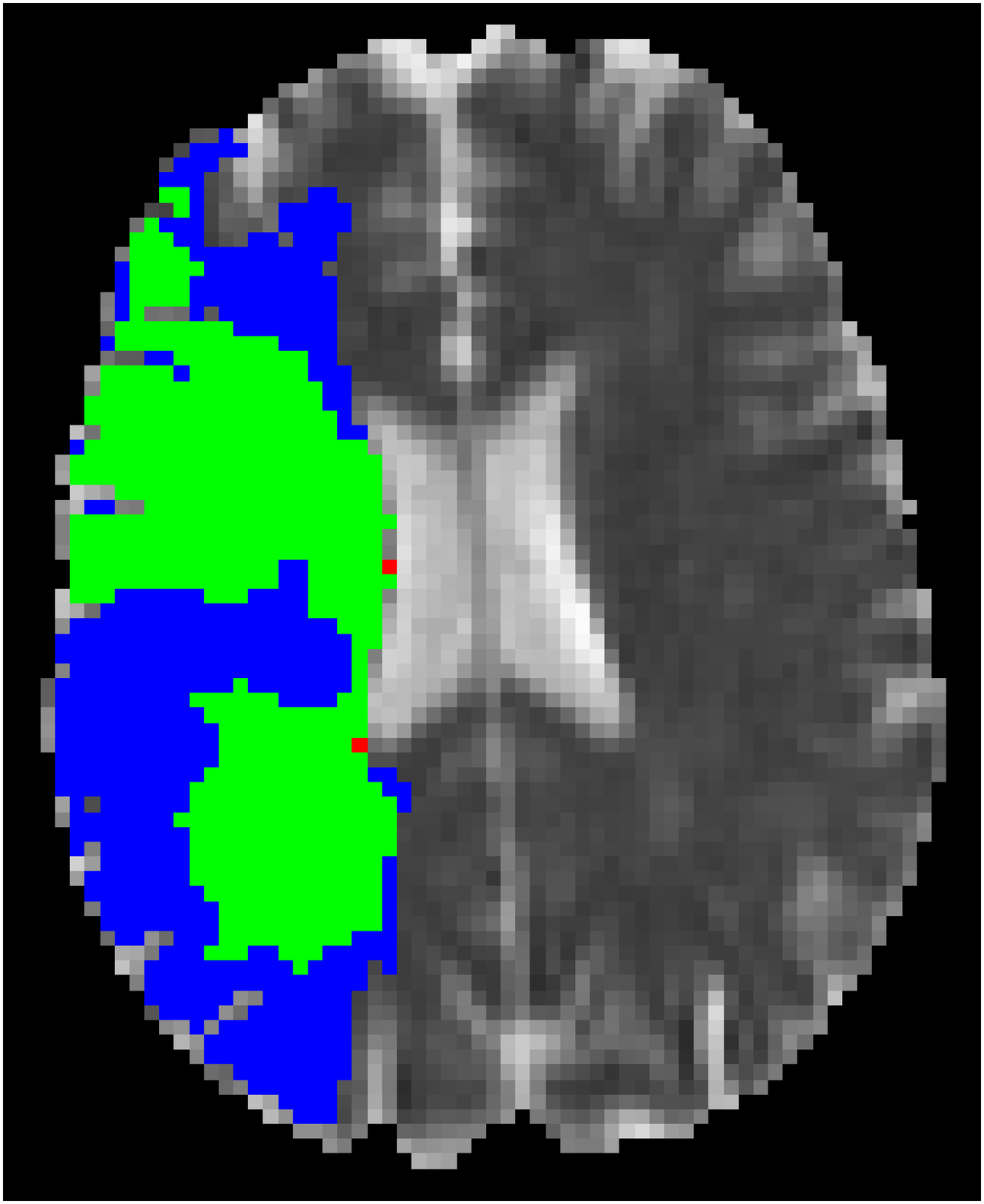}
		\caption{Case 15}
	\end{subfigure} 
	\quad
	\begin{subfigure}[b]{0.2\textwidth}
		\includegraphics[width=\textwidth,trim={0px 0px 0px 0px},clip]{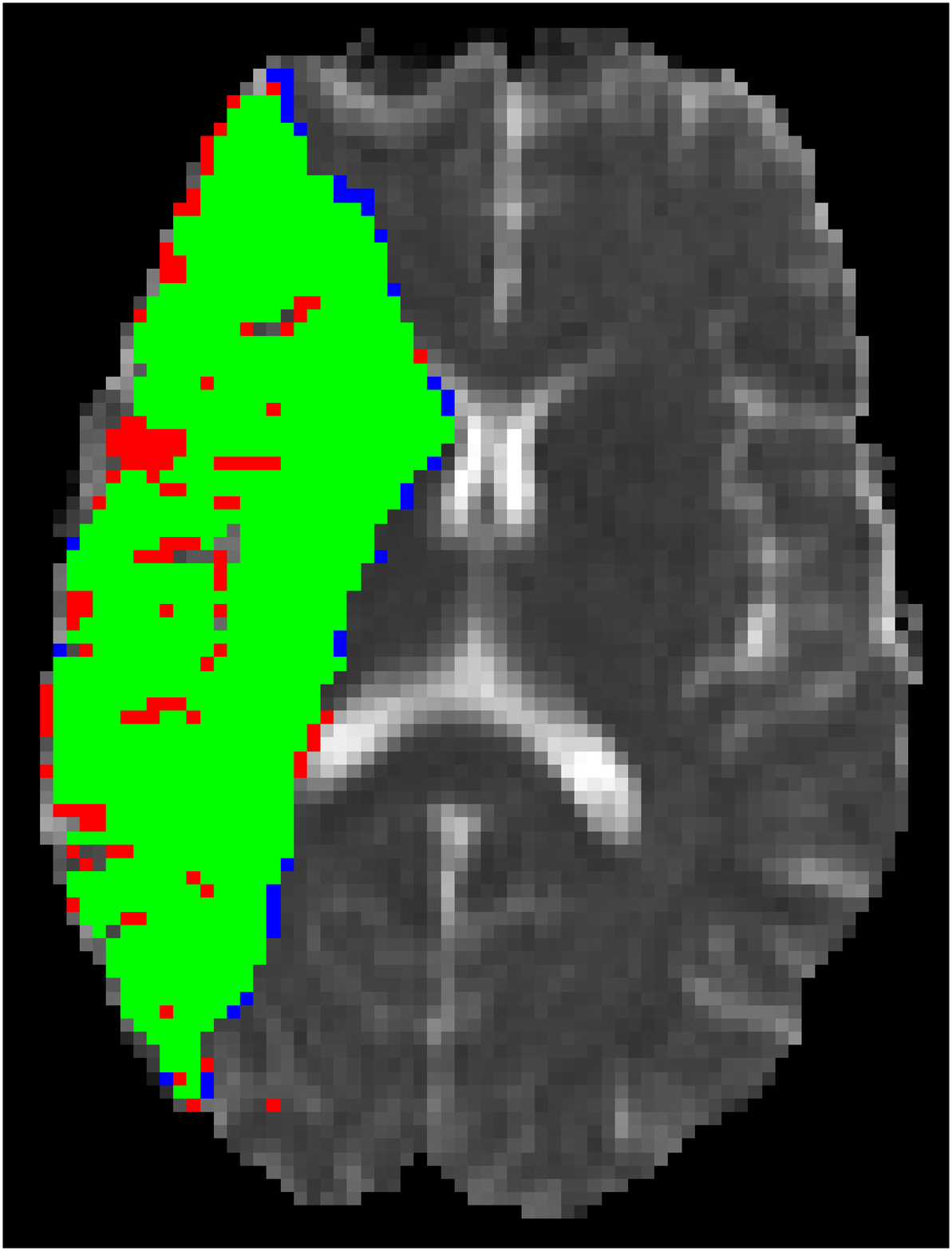}
		\caption{Case 26}
	\end{subfigure}
	\caption{Output segmentation masks of representative cases from the training images of ISLES 2015 SPES dataset. On all images, true positives are denoted in green, false positives in red and false negatives in blue.}
	\label{fig:qspes}
\end{figure*}

Figure~\ref{fig:qspes} shows qualitative results of four representative segmentation examples  from the proposed method. In general, the majority of the lesion is correctly segmented with minor border and small hole inaccuracies as seen in cases 11 and 26. Other less typical errors include under or oversegmentation of the lesion, as seen in case 2 where false positives are found on the upper part of the lesion. In the example of case 15, the lesion is undersegmented due to a confounding unusual appearance of some parts.

\subsection{Blind challenge evaluation}

To compare the proposed methodology against other state-of-the-art methods for acute stroke we submit our final approach for blind external evaluation in the ISLES 2015 challenge framework. The web platform used to hold the 2015 ISLES workshop \cite{isles15website} remains open for later submission and maintains an ongoing challenge leaderboard where the average testing set results are publicly displayed. Since the gold standard is hidden for the testing images, a fair and direct method comparison is possible. For evaluation in the challenge framework of ISLES 2015, we use the four networks trained for the Proposed approach during the cross-validation experiment, one from each fold, and average their outputs to produce a single testing patch prediction. The testing images are then segmented as described in Section~\ref{subsec:segmentation} using the $T_h$ and $S_{min}$ set in the cross-validation experiment. In this way, the challenge results are produced with the same networks trained in the cross-validation experiment.

\subsubsection{Challenge results}

Tables~\ref{table:siss}~and~\ref{table:spes} shows the top five entries as ranked by DSC of the ongoing testing leaderboard results for the SISS and SPES sub-tasks respectively. The proposed methodology achieves state-of-the-art performance in both sub-tasks, ranking first out of 74 entries in the SISS leaderboard and first out of 41 entries in the SPES leaderboard. 
As compared with the next best entries, we achieve similar or higher DSC with a 12\% and 28\% lower Haussdorf distance in the SISS and SPES sub-tasks respectively. Additionally, in the SPES dataset we also obtain an 8\% higher sensitivity.

\begin{table*}[t]
	\centering
	\caption{Top 5 out of 74 entries of the ongoing SISS testing leaderboard \cite{isles15website} as ranked by average DSC.}
	\begin{tabular}{rlrrrr}
		\hline
		Rank & User & DSC & PPV & Sensitivity & HD \\ 
		\hline
		\textbf{1} & \textbf{clera2} \textbf{(ours)} & \textbf{0.59} $\pm$ \textbf{0.31} & \textbf{0.65} $\pm$ \textbf{0.35} & \textbf{0.60} $\pm$ \textbf{0.30} & \textbf{34.7} $\pm$ \textbf{28.9} \\
		2 & kamnk1 \cite{kamnk1} & 0.59 $\pm$ 0.31 & 0.68 $\pm$ 0.33 & 0.60 $\pm$ 0.27 & 39.6 $\pm$ 30.7 \\
		3 & zhanr6 \cite{Zhang2018} & 0.58 $\pm$ 0.31 & 0.60 $\pm$ 0.33 & 0.68 $\pm$ 0.24 & 38.9 $\pm$ 35.3 \\
		4 & lianl1 & 0.57 $\pm$ 0.29 & 0.58 $\pm$ 0.30 & 0.64 $\pm$ 0.29 & 43.0 $\pm$ 30.5 \\
		5 & saliz1 & 0.57 $\pm$ 0.31 & 0.54 $\pm$ 0.31 & 0.67 $\pm$ 0.29 & 41.1 $\pm$ 36.7 \\
		\hline
	\end{tabular}
	\label{table:siss}
\end{table*}

\begin{table*}[t]
	\centering
	\caption{Top 5 entries out of 41 of the ongoing SPES testing leaderboard \cite{isles15website} as ranked by average DSC.}
	\begin{tabular}{rlrrrr}
		\hline
		Rank & User & DSC & PPV & Sensitivity & HD \\ 
		\hline
		\textbf{1} & \textbf{clera2} \textbf{(ours)} & \textbf{0.84} $\pm$ \textbf{0.10} & \textbf{0.82} $\pm$ \textbf{0.15} & \textbf{0.89} $\pm$ \textbf{0.06} & \textbf{20.7} $\pm$ \textbf{13.9} \\
		2 & mckir1 \cite{mckir}& 0.82 $\pm$ 0.08 & 0.83 $\pm$ 0.10 & 0.82 $\pm$ 0.14 & 29.0 $\pm$ 16.3 \\
		3 & cheng5 & 0.81 $\pm$ 0.11 & 0.81 $\pm$ 0.12 & 0.81 $\pm$ 0.14 & 22.7 $\pm$ 12.6 \\
		4 & maieo1 \cite{maieo}& 0.81 $\pm$ 0.09 & 0.84 $\pm$ 0.08 & 0.80 $\pm$ 0.14 & 23.6 $\pm$ 13.0 \\
		5 & ghosp1 & 0.80 $\pm$ 0.11 & 0.80 $\pm$ 0.15 & 0.83 $\pm$ 0.11 & 57.1 $\pm$ 25.4 \\
		\hline
	\end{tabular}
	\label{table:spes}
\end{table*}

\section{Discussion}

We have performed both qualitative and quantitative evaluations of the proposed methodology in two different tasks without any dataset specific tuning of training hyper-parameters. The methodology has been shown to perform equally well for the acute or sub-acute stages and with different combinations of MRI modalities. The results are improved with respect to the Baseline thanks to the combined approach to alleviate data imbalance and also through the explicit learning of features based on the brain symmetry. Additionally, the method is fast in inference, taking under 3 minutes to pre-process and predict a new image. 

The proposed methodology demonstrates state-of-the-art performance ranking 1st by average DSC while having a smaller HD as compared with the next best method in both challenges. Moreover, we are the first U-Net based approach in the online testing leaderboard to outperform the best 2015 ISLES workshop entries. In the SISS sub-task, we obtain a similar DSC but with lower HD than the next best method. In contrast with the approach by Kamnitsas et al \cite{kamnk1}, we can avoid the use of the additional post-processing step with conditional random fields that needs several image dependent configurable parameters. In our case, the use of a U-Net based architecture that provides whole patch predictions allows performing highly overlapped segmentations without a large increase of inference time or introducing additional configurable parameters. In the SPES sub-task, we obtain a higher sensitivity with a lower HD as compared with the next best method by McKinley et al \cite{mckir} that used a random decision forest classifier with several hand-crafted features over $3\times3\times3$ and $5\times5\times5$ neighbourhoods including local texture features, mean intensity, skewness, etc. By using a deep learning based method, the feature representation is learned at training time without having to rely on manually testing and finding the most appropriate ones for each specific task. Despite the good relative performance, the qualitative results show that the proposed methodology is still limited by inaccurate borders, missing lesion parts and other confounding factors. Furthermore, while the found minimum lesion size maximize the desired metrics along all training images they might still filter out some small lesions at testing time.
 
\section{Conclusions}

In this work, we have presented a methodology that achieves state-of-the-art performance in two different stroke lesion segmentation tasks. To the best of our knowledge, the proposed methodology is the first to obtain competitive results in both the ISLES 2015 SISS and SPES sub-tasks with the same approach. We have achieved these results by doing both regularization of the training procedure and providing additional meaningful information for lesion segmentation. Useful features using the brain symmetry could not be learned as the employed patch size is too small to include both hemispheres. The proposed symmetric modality augmentation facilitates using the similarity between hemispheres to improve lesion localization without using larger patches that would worsen class imbalance. The use of additional informative modalities can be generalized with different augmentation techniques to facilitate the learning of more discriminative and meaningful features for the task at hand. Moreover, we have shown the big influence class imbalance can have in reducing distant outliers and false positives that provide a lower Hausdorff distance at testing time. By using a combined approach we achieve a less biased segmentation with a better balance between sensitivity and specificity. In the clinical setting, deep learning based methods can additionally benefit from related techniques such as transfer learning to learn an unrelated task with a lack of training examples \cite{Pang2017} or to perform domain adaptation with few images \cite{Valverde2019}. The proposed methodology is made publicly available for the scientific community \cite{githubrepo}.

\section*{Acknowledgements}

Jose Bernal holds an FI-DGR2017 grant from the Catalan Government with reference number 2017FI\_B00476. This work has been partially supported by Retos de Investigaci\'on  TIN2015-73563-JIN and DPI2017-86696-R from the Ministerio de Ciencia, Innovaci\'on y Universidades. The authors gratefully acknowledge the support of the NVIDIA Corporation with their donation of the TITAN X GPU used in this research.







\end{document}